\documentclass{article}

\usepackage[preprint,nonatbib]{neurips_2020}

\usepackage[utf8]{inputenc} 
\usepackage[T1]{fontenc}    
\usepackage{hyperref}       
\usepackage{url}            
\usepackage{booktabs}       
\usepackage{amsfonts}       
\usepackage{nicefrac}       
\usepackage{microtype}      

\usepackage{lipsum}
\usepackage{amsmath}
\usepackage{enumitem}
\usepackage{graphicx}
\usepackage{subcaption}
\usepackage{multirow}
\usepackage{xcolor}

\newcommand{\fig}[1]{Fig.\ \ref{fig:#1}}

\providecommand{\smallsec}[1]{{\vspace{0em}\textbf{#1}}}

\title{Empirically Verifying Hypotheses Using Reinforcement Learning}

\author{%
  Kenneth Marino \\
  Carnegie Mellon University\\
  \texttt{kdmarino@cs.cmu.edu} \\
  \And
  Rob Fergus \\
  New York University \\
  \texttt{fergus@cs.nyu.edu} \\
  \And
  Arthur Szlam \\
  Facebook Artificial Intelligence Research \\
  \texttt{aszlam@fb.com} \\
  \And
  Abhinav Gupta \\
  Carnegie Mellon University\\
  Facebook Artificial Intelligence Research \\
  \texttt{abhinavg@cs.cmu.edu} \\
}

\begin{document}

\maketitle

\begin{abstract}
This paper formulates hypothesis verification as an RL problem. Specifically, we aim to build an agent that, given a hypothesis about the dynamics of the world, can take actions to generate observations which can help predict whether the hypothesis is true or false. Existing RL algorithms fail to solve this task, even for simple environments. 
In order to train the agents, we exploit the underlying structure of many hypotheses, factorizing them as \{{\em pre-condition}, {\em action sequence}, {\em post-condition}\} triplets. By leveraging this structure we show that RL agents are able to succeed at the task. Furthermore, subsequent fine-tuning of the policies allows the agent to correctly verify hypotheses not amenable to the above factorization.
\end{abstract}

\section{Introduction}
Empirical research on early learning \cite{Gopnik12,kushnir2005young} has shown that infants build an understanding of the world around by constantly formulating hypotheses about how some physical aspect of the world might work and then proving or disproving them through deliberate play. Through this process the child builds up 
a consistent causal understanding of the world. This contrasts with manner in which current ML systems operate. Both traditional i.i.d and interactive learning settings use a single user-specified objective function that codifies a high-level task, and the optimization routine finds the set of parameters (weights) which maximizes performance on the task. The learned representation (knowledge of how the world works) is embedded in the weights of the model - which makes it harder to inspect, hypothesize or even enforce domain constraints that might exist. On the other hand, hypothesis generation and testing is a process explored in classical approaches to AI \cite{Brachman04}. In this paper we take a modest step towards the classical AI problem of building an agent capable of testing hypotheses about its world using modern statistical ML approaches.

\begin{figure}[t]
\begin{center}
\includegraphics[width=0.62\textwidth]{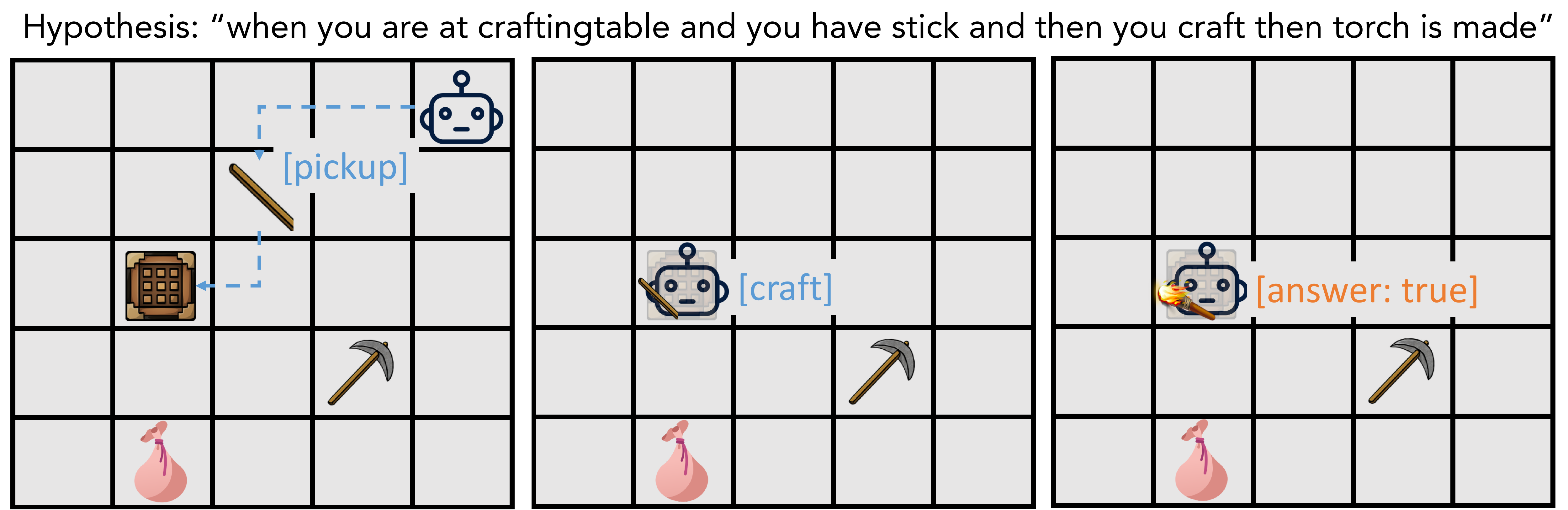}
\end{center}
\definecolor{orange}{rgb}{0.93, 0.53, 0.18}
\definecolor{blue2}{rgb}{0.19, 0.55, 0.91}
\caption{Example of a "Crafting" world. The agent verifies a hypothesis (provided a text) about a causal relationship. Acting according a \textcolor{blue2}{learned policy}, the agent manipulates the observation to one that allows a learned \textcolor{orange}{predictor} to determine if the hypothesis is true. The learning of policy and predictor is aided by a pretraining phase, during which an intermediate reward signal is provided by utilizing hypotheses that factor into \{{\em pre-condition state}, {\em action sequence}, {\em post-condition state}\}. 
}
\label{fig:teaser}
\end{figure}

The problem we address is illustrated in Figure~\ref{fig:teaser}. Agents are placed in a world which has several interactive elements. They are provided with a hypothesis (an "action sentence" \cite{pearl2009causality}) about the underlying mechanics of the world via a text string
(e.g. "$\mathcal{A}$ will be true if we do $\mathcal{B}$"). The task is to determine if the hypothesis is true or not. This problem cannot be solved without {\em interaction} with a dynamic world (comparing the state before and after taking action $\mathcal{B}$).

A key novelty in our work is formulating the task in a manner that permits the application of modern RL methods, allowing raw state observations to be used rather than abstract Boolean expressions of events. To do this, we use a model composed of two different deep parametric functions which are learned through interaction: (i) a policy that generates observations relevant to verification of the hypothesis and (ii) a prediction function which uses the observations to predict whether it is true.

We first show that even in simple environments, agents trained end-to-end using deep reinforcement learning methods cannot learn policies that can generate observations to verify the hypothesis. To remedy this, we exploit the underlying structure of hypotheses -- they can often be formulated as a triplet of a pre-condition ($\mathcal{P}$), an action sequence (collectively $\mathcal{B}$), and a post-condition ($\mathcal{A}$) that is causally related to the pre-condition and actions. Using this common structure, we are able to seed our action policy to learn behaviors which alter the truth of the pre-condition and post-condition. 
This allows agents to learn policies that can generate meaningful observations for training the prediction function. We further show that these policies can be adapted to learn to verify more general hypotheses that do not necessarily fit into the triplet structure.

\section{Related Work}
{\bf Knowledge representation and reasoning (KRR)} \cite{Brachman04} is a central theme of traditional AI. Commonsense reasoning \cite{Davis90,Davis2015,liu2004conceptnet} approaches, e.g. CYC \cite{Lenat95}, codify everyday knowledge into a schema that permits inference and question answering. However, the underlying operations are logic-based and occur purely within the structured representation, having no mechanism for interaction with an external world. Expert systems \cite{giarratano1998expert} instead focus on narrow domains of knowledge, but are similarly self-contained. Logic-based planning methods \cite{fikes1971strips,Colaco15} generate abstract plans that could be regarded as action sequences for an agent. By contrast, our approach is statistical in nature, relying on Reinforcement Learning (RL) to guide the agent. 

Our approach builds on the recent interest \cite{mao2018,garcez2012neural} in neural-symbolic approaches that combine neural networks with symbolic representations. In particular, some recent works \cite{Zhang15,Lu18} have attempted to combine RL with KRR, for tasks such as navigation and dialogue. These take the world dynamics learned by RL and make them usable in declarative form within the knowledge base, which is then used to improve the underlying RL policy. In contrast, in our approach, the role of RL is to verify a formal statement about the world. Our work also shares some similarity with \cite{konidaris2018skills}, where ML methods are used to learn mappings from world states to representations a planner can use.

{\bf Causality and RL:} There are now extensive and sophisticated formalizations of (statistical) causality \cite{pearl2009causality}. These provide a framework for an agent to draw conclusions about its world, and verify hypothesis as in this work. This is the approach taken in \cite{dasgupta2019causal}, where RL is used to train an agent that operates directly on a causal Bayesian network (CBN) in order to predict the results of interventions on the values on its nodes. 

In contrast, the approach in this work is to sidestep this formalization with the hope of training agents who test hypotheses without building explicit CBNs. Unlike \cite{dasgupta2019causal}, our agents intervene on the actual world (where interventions may take many actions), rather than the abstract CBN. Nevertheless, we find that it is necessary to add inductive bias to the training of the agent; here we use the pretraining on ($\mathcal{P}$, $\mathcal{B}$, $\mathcal{A}$) triplets. These approaches are complementary; one could combine explicit generation and analysis of CBNs as an abstract representation of an environment with our training protocols.

Our work is thus most similar to \cite{denil2016learning}, which uses reinforcement learning directly on the world, and the agent gets reward for answering questions that require experimentation. However, in that work (and in \cite{dasgupta2019causal}), the ``question'' in each world is the same; and thus while learning to interact led to higher answer accuracy, random experimental policies could still find correct answers. On the other hand, in this work, the space of questions possible for any given world is combinatorial, and random experimentation (and indeed vanilla reinforcement learning) is insufficient to answer questions.

{\bf Cognitive development:} Empirical research on early learning \cite{Gopnik12,kushnir2005young} shows infants build an understanding of the world in ways that parallel the scientific process: constantly formulating hypotheses about how some physical aspect of the world might work and then proving or disproving them through deliberate play. Through this process the child builds up an abstract consistent causal understanding of the world. Violations of this understanding elicit surprise that can be measured \cite{spelke1992origins}. 

{\bf Automated Knowledge Base completion:} This work is also related to knowledge base completion \cite{Fader11,Bordes13,Suchanek07}, especially as formulated in \cite{riedel2013relation}. Instead of using other facts in the knowledge base or a text corpus to predict edges in the KB, here the agent needs to act in a world and observe the results of those actions. This recalls \cite{Mitchell18}, where the system verifies facts it has previously hypothesized by searching for corroboration in the corpus.

{\bf Automation of the scientific process:} has been tried in several domains. 
Robotic exploration of chemical reactivity was demonstrated \cite{granda2018} using ML techniques. 
\cite{king2009automation} developed a robot scientist that explored geonomics hypotheses about yeast and experimentally tested them using laboratory automation. In biochemistry \cite{vanlier2014optimal} used Bayesian methods for optimal experiment design. More generally, the Automated Statistician project~\cite{AutoStat} uses a Bayesian approach to reason about different hypotheses for explaining the data, with the aim of creating interpretable knowledge. 

{\bf Embodied Question and Answering:} The problem studied in this paper is closely related to the embodied visual question-answering problem in \cite{das2018embodied}. Indeed, our basic formulation is a particular case of the most general formulation of embodied QA, as the agent is rewarded for successfully answering questions about the world that require interaction. However, the form of the questions is different than those considered in that work, as they may require drawing a conclusion about the {\it dynamics} of the world, rather than a static property. Even the questions about static properties we are interested in have a different flavor, as they encode rules, rather than statements about the current configuration. Our approach is built around hypothesis-conclusion structure special to these questions. There is also a large body of work on visual QA \cite{kafle2017visual,Wu16} and text-based QA \cite{Rajpurkar18}. From this, most relevant to our work is \cite{Wu_2016_CVPR} who use a structured knowledge base to augment standard QA techniques.


\section{The Hypothesis Verification Problem}

An agent is spawned in a world sampled from a distribution over possible worlds. In the case of ``Crafting'', shown in Figure~\ref{fig:teaser}, there are items lying around that the agent can pick up and combine using a ``craft'' action. The exact dynamics change for every newly instantiated world; so in one world, taking a craft action with a stick might produce a torch, and in another, it might produce a pickaxe. At the start of each episode, the agent is given a hypothesis about the world, such as the one shown at the top of Figure~\ref{fig:teaser}. The agent gets a reward when it correctly answers if that hypothesis is true or false. Because the dynamics and rules change each episode, the agent must learn to interact with the world in order to decide if the hypothesis is true. In Figure~\ref{fig:teaser} the agent picks up the stick and does a craft action to see that a torch is created. It then has enough information to decide the hypothesis is true, and the agent receives reward for verifying the hypothesis correctly.

In this work, we will structure our hypotheses using templated language. One could imagine using more expansive formal symbolic systems (e.g.~first order logic), or alternatively, using natural language descriptions of the hypotheses. The former might allow interfacing with symbolic solvers or otherwise using combinatorial approaches; whereas the latter would allow scaling annotation to untrained humans. We choose templated language because it is simple, and sufficient for the environments on which we test, which are already challenging for standard RL. Moreover, in our view it is a good starting point for further work that would use either more sophisticated formal representations or more natural language representations.

{\bf Formal Definition}
\label{section:formaldefinition}
We first define a world as a set of states and actions with Markovian dynamics (an MDP without a reward). We define an environment $\mathcal{E}$ as a distribution over a set of worlds $\mathcal{W}$ and hypotheses $\mathcal{H}$.
A world $W\in \mathcal{W}$ is specified by rules $L_W$ which describe the dynamics of the world. We can define this reward-less MDP of one specific world $W$ as $MDP_W = \{S_W, A_W, T_W\}$ where the state space $S_W$ includes the position and state of objects in the world (e.g. the placement of the agents and the object), $A_W$ is the action space for that environment, and $T_W$ is the transition function. Note that $T_W$ depends on $L_W$, the rules of this specific world. Actions have different consequences depending on $L_W$. 

Now $\mathcal{E}$ is an episodic POMDP where each episode consists of sampling\footnote{See Appendix~\ref{appendix:templates} for details on sampling procedures} a $W$ and $h$. (
$G$ is a ground-truth function that takes in the hypothesis $h$ and world $\mathcal{W}$ and outputs \{true, false\}. In this work, hypotheses are generated via templated language and their truth function $G$ depends on $W$, more specifically $L_W$. The episode ends when the agent executes either the $\mathsf{true}$ or $\mathsf{false}$ action. 

Given a world $W$ and hypothesis $h$, an agent gets reward: \\ \\
$R_{Hyp} = \left \{ \begin{array}{ll}
        +1 & a = G(h, W) \\
        -1 & a = \neg G(h, W) \\
        0 & otherwise \\
    \end{array} \right.
$

The observation in this POMDP is $o = (s_W, h)$, the state from the world $W$ plus the hypothesis. The state is $s = (s_W, h, L_W)$. This includes the rule $L_W$ which is not visible in the observation. The action space is just $A_W \cup \{\mathsf{true}, \mathsf{false}\}$ for any $W$ (they are the same for a given environment); and $T = T_W$.
Note that the transition function $T$ depends on the (hidden) $L_W$. The goal of hypothesis verification is now to discover the truth of $h$, which depends on $L_W$.  
 
\section{Methodology}
\label{section:method}
{\bf RL baseline}
Given the formulation of the hypothesis verification problem as a POMDP, we could try to solve it using an RL agent with $a = \pi(O_t, h)$, where $O_t=\{o_t, o_{t-1},\ldots o_{t-K}\}$. Here $o$ is an observation of the current world, $K$ is a history window size, and $h$ is the hypothesis for the current world. We found that standard RL agents struggle to solve this problem.

To make the problem easier to solve, we can augment the reward with direct supervision of a prediction network
$f(O_t, h)$ which takes in the last $K$ observed observations of the environment and the hypothesis and predicts whether or not the hypothesis is true. Our policy network now, instead of taking $\mathsf{true}$ or $\mathsf{false}$ actions, takes a special $\mathsf{stop}$ action which is replaced by $\mathsf{true}$ or $\mathsf{false}$ based on the prediction of $f$.

Even with the augmented supervision, as shown in \fig{stage280maxstd}, an RL baseline is not able to solve the task. In order determine whether a hypothesis is true, the agent needs to take the correct sequence of actions related to the hypothesis. But in order to know that a particular sequence of actions was the right one, it needs to be able to correctly predict the hypothesis. Guessing with no information gives a zero average reward, and despite the supervision on the output of the predictor, the predictor does not see state transitions that allow it to learn.

{\bf Pretraining using Triplet Hypotheses}
\label{section:tripletpretrain}
In light of the difficulties directly training an RL model using terminal reward and hypothesis prediction supervision, we take advantage of the fact that many causal statements about the world have the form: $(\text{\it pre-condition}, \text{\it action sequence})\implies \text{\it post-condition}$

We define this formally in Appendix~\ref{appendix:tripletdef}, but informally this means that when the state meets a ``pre-condition'' and an ``action sequence'' is taken, this will result in the state meeting a ``post-condition.'' In Figure~\ref{fig:teaser} the pre-condition is having a stick and being at craftingtable, the action sequence is craft, and the post-condition is that a torch is made.

This structure can be converted into a reward function that can be used to pretrain the agent policy $\pi$. The idea is to reward the agent for taking actions which alter the truth of the pre-condition and post-condition (i.e. changing the world state so that pre/post-conditions are met or not). If it matches the pre-condition state and takes the action, if the statement is true, the post-condition should toggle from false to true in the world. Similarly, if post-condition changes but the pre-condition did not change, the statement must be false. This can be formalized in the following reward function to encourage the agent to toggle pre-condition and post-condition states:

$R_{\mathsf{pre}} = \left \{ \begin{array}{ll}
        +C & a = \mathsf{stop} \ \& \  \\
        & \text{pre-condition changed in last $K$ steps}\\
        0 & otherwise \\
    \end{array} \right.$ 
    
$R_{\mathsf{pre+post}} = \left \{ \begin{array}{ll}
        +C & a = \mathsf{stop} \ \& \ \text{post+pre-condition} \\
        & \text{changed in last $K$ steps} \\
        0 & otherwise \\
    \end{array} \right.$ 
    
This encourages the policy $\pi$ to change the pre-condition and post-conditions (via pre-condition) in the last $K$ steps, so that a predictor looking at the last $K$ observations will be able to deduce the truth value of the hypothesis. More generally, training with this reward function forces the policy network to ground text concepts (e.g. the text "stick" means [object\_stick]) and also captures the causal rules within the world. Consequently, following pretraining, the policy network can then be fine-tuned using the original reward function $R_\mathsf{Hyp}$. Since the policy network is no longer random, a robust prediction network $f$ can also be learned. While not all hypotheses fit into the triplet format, we show in the experiments that the knowledge captured by the policy and prediction networks during this phase of training can generalize to less structured hypotheses. We have a set of hypotheses for each world that contains only these triplet-structured hypotheses. We use this set for our pretraining.

{\bf Training using Triplet hypothesis}
\label{sec:train_trip}
After pretraining the policy $\pi$, we further train $\pi$, as well as the prediction network $f$ using the same set of triplet hypotheses, but now using $R_{\mathsf{Hyp}}$ instead of $R_{\mathsf{pre}}$ or $R_{\mathsf{pre+post}}$. Two variants are explored: (i) ``fixed'' --  keep $\pi$ fixed but train the prediction network $f$ and (ii) ``finetune" -- finetune $\pi$ and train $f$. Performance in this phase is used to select promising models for subsequent stages of training. Specifically, runs achieving <90\% accuracy (see the appendix for alternate cutoff) are eliminated.

{\bf Adaptation to non-triplet hypotheses}
\label{sec:train_adapt}
Next, we want to show that we can adapt our networks to hypotheses other than those that fall neatly into the triplet structure. To adapt to the larger set of hypotheses, we start with the networks trained previously on triplet templates. During this training stage, the triplet-form constraint is relaxed and training proceeds with both triplet and non-triplet hypotheses (see Sec.~\ref{section:hypothesisconstruct}), using an even split between the two types. 

\section{Evaluation Environments}
In this section we describe 
our environments
The environments $\mathcal{E}$ are designed so that the prior probability $p(h=true)=0.5$ and the initial observation $o_0$ does not contain information about $h$.

{\bf Environments}
We created four different environments for hypothesis verification. ColorSwitch, Pushblock and Crafting are all gridworld-based environments. A fourth enviornment is created by adapting the standard Cartpole task to include interactive elements. See Fig. 2.

{\noindent \bf ColorSwitch}: The agent is placed in a world with one or more color switches which are randomly either ``on'' or ``off'' and a door which is either open or closed. The agent is able to move and toggle the switch positions. One of the switches in the world, when in the correct position (can be either on or off) will cause the door to open. The other switches have no effect. Hypotheses in this environment relate to the color and position of switches and how that opens or closes the door.

{\noindent \bf Pushblock}: The agent is placed in a world with a block which can be pushed by the agent, and a door. The agent can move and push on the block. The door opens when the block is in a particular part of the grid: ``up'' -- top two rows, ``down'' -- bottom two rows, ``left'' -- leftmost two rows, ``right'' -- rightmost two rows. The hypotheses in this environment relate to the position of the pushblock and how that affects the door.

{\noindent \bf Crafting}: The agent is placed in a world with crafting rules similar to that of the popular Minecraft game. The agent is spawned along with a number of crafting items, and a crafting location. The agent is able to move, pick up items into its inventory and use the crafting location using special crafting actions. There is some true ``recipe'' which produces some new item in the agent's inventory.

{\noindent \bf Cartpole}: This is the standard classic control cartpole problem where a pole is attached by an un-actuated joint to a cart. The regular actions are ``left'' and ``right'' and if the pole falls, the episode ends. In our modification, there are ``zones'' (an interval on the x axis) where the physical laws of the cartpole change by either changing the gravity constant, or applying a ``wind force'' blowing the cart in one direction. Like in ColorSwitch, the zones are specified by color. Typically one color zone has an effect and the other is a decoy zone that has no effect. The hypotheses relate to which color zones correspond to what changes to the physics. 

\begin{figure}[h]
\begin{center}
   \includegraphics[width=0.8\linewidth]{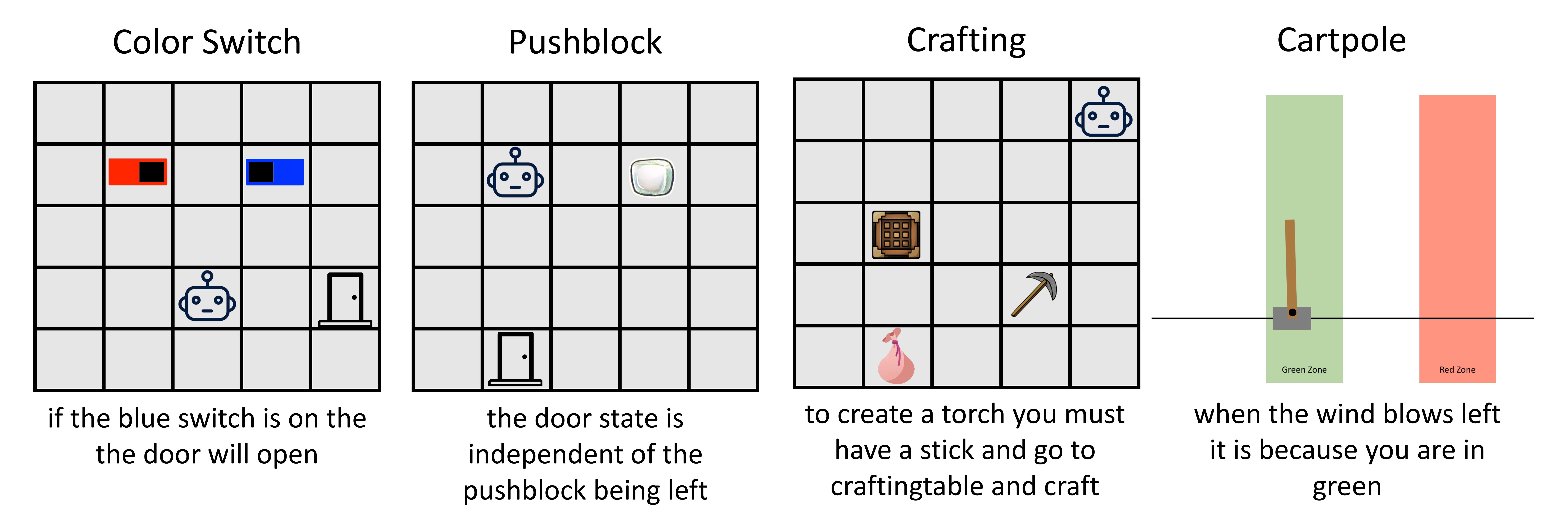}
\end{center}
\caption{Examples of the four environments used in our experiements: ColorSwitch, Pushblock, Crafting and Cartpole.}
\label{fig:environments}
\end{figure}

In the grid world environments, items are randomly generated in a $5$ by $5$ grid. The world observation is given by a 1-hot vector for each grid location and inventory. The hypothesis is encoded as sequence of tokens. In Cartpole the state is the standard dynamics as well as a 1-hot vector specifying the location and color of the zones.

{\bf Hypothesis Construction}
\label{section:hypothesisconstruct}
We now describe how the hypotheses for each world in each environment are automatically generated via templates. Three different varieties are considered: (i) triplet hypotheses, (ii) general templates and (iii) special case templates. (See Appendix~\ref{appendix:rawtemplates} for all of the possible templates for an environment and further details).

{\noindent \bf Triplet hypotheses}:
\label{triplethyp}
Here the hypotheses takes the form of an explicit logical statement: 
(pre-condition, action sequence) $\implies$ post-condition. When the pre-condition is true, and the action sequence is performed, the post-condition will become true. To generate triplet hypotheses, we: (i) randomly select a pre-condition template from a set list; (ii) randomly select an action template; (iii) randomly select a post-condition template; and (iv) fill in any entities in the final template. In our example from Fig.~\ref{fig:teaser} this would be (``when you are at crafting table and you have stick''; ``and then you craft''; ``then torch is made'').

{\noindent \bf General templates}:
Instead of drawing a template from the triplet form, a single template for the hypothesis is drawn and the values populated. For instance, in Pushblock, a template might be ``the door can only be opened when the pushblock is PUSHBLOCK\_POSITION''
and then ``left'' would be drawn for PUSHBLOCK\_POSITION. These templates are more general than the triplet ones in that they have no explicit (pre-condition, action sequence and post-condition) structure.

{\noindent \bf Special cases}:
\label{section:specialcases}
We also use more difficult and general hypothesis templates. These cannot be neatly fit into a triplet format by rewording, and may not fully describe the rules of the world. Some examples of these harder templates are: (i) Negating effects (e.g. ``door is not open"); (ii) Negating conditions (e.g. ``switch is not on"); and independence (e.g. ``door independent of blue switch"). 

\section{Experiments}
\begin{figure}[h]
\begin{center}
    \includegraphics[width=1\linewidth]{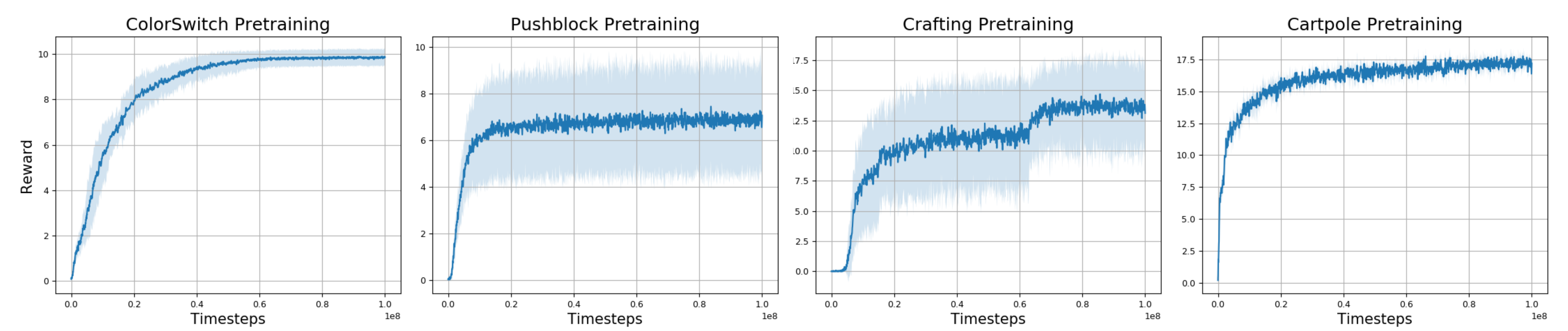}
\end{center}
\caption{Pretraining reward ($R_{\mathsf{pre}}+R_{\mathsf{pre+post}}$) on our four environments.}. 
\label{fig:pretrainresults}
\end{figure}

Figure~\ref{fig:pretrainresults} shows results from learning with pretraining rewards $R_{\mathsf{pre+post}}$ and $R_{\mathsf{post}}$. There is relatively little variance, with all runs achieving near the theoretical maximal rewards\footnote{For Pushblock, sometimes the block can be stuck against a wall, so not all worlds are solvable} 
For Crafting and Cartpole, $R_{\mathsf{pre+post}}$ is not always achievable if true and distractor items are far away from each other. See Appendix~\ref{appendix:pretrainingdisc} for further discussion. \footnote{For the gridworld environments we show variance runs on $25$ random seeds and $5$ for Cartpole. We provide further hyperparameters, and training details in Appendix~\ref{appendix:learningparams} and ~\ref{appendix:networkparams} as well as network architecture and details in Appendix~\ref{appendix:networkparams}}.

In Figure~\ref{fig:stage280maxstd}, we show the results on non-triplet adaptation (Sec.~\ref{sec:train_adapt}). As discussed in Section~\ref{section:hypothesisconstruct}, this stage includes the more difficult, non-triplet templates not seen during pretraining or during triplet hypothesis training. We also break down the final hypothesis prediction accuracy for our methods in Table~\ref{table:success}. This allows us to see whether our methods were able to adapt to non-triplet hypotheses. All network architectures for the policy and prediction networks, as well as hyper-parameters are the same for all methods.

\smallsec{RL baseline} Figure~\ref{fig:stage280maxstd} shows the RL baseline at chance-level performance, the only exception being CartPole were it achieves $\sim60\%$, versus the $\sim90\%$ of our approaches.  Note that this is the RL baseline {\it with} the supervised predictor, as discussed in Section~\ref{section:method}. This poor performance is a result of training both policy and predictor from scratch; see Figure \ref{fig:oracletrain}.  

\begin{figure}[h]
\begin{center}
    \includegraphics[width=1\linewidth]{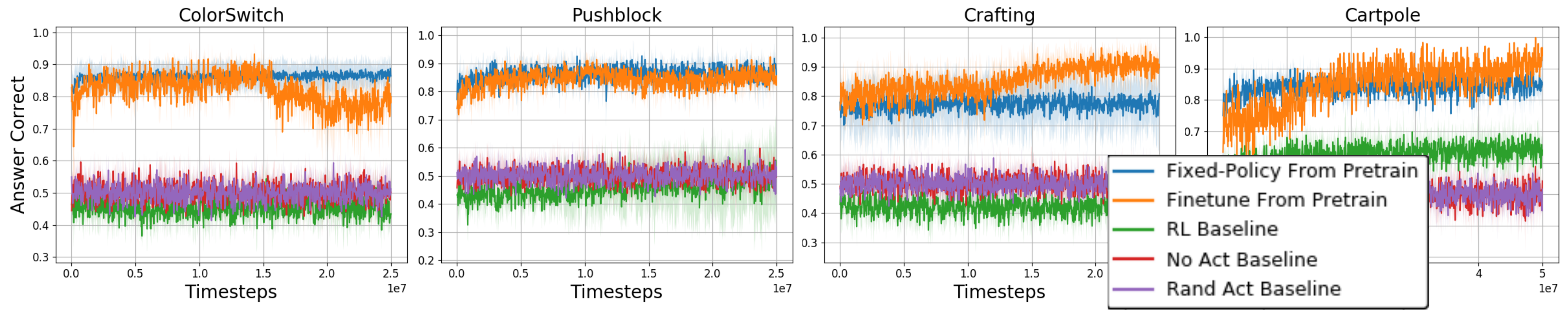}
\end{center}
\caption{Hypothesis prediction accuracy on {\it both} triplet {\it and} non-triplet hypotheses (1-1 ratio) for the ColorSwitch, Pushblock, Crafting and Cartpole environments, using $R_\mathsf{Hyp}$ reward for training.}
\label{fig:stage280maxstd}
\end{figure}

\smallsec{Other baselines}
We also include two other simple baselines ``no act'' and ``random act.'' The ``no act" baseline simply takes the $\mathsf{stop}$ action at $t=0$, forcing the prediction network to give an answer from just the first observation. This fails because the agent needs to take actions in the world to be able to predict the hypothesis accurately. This also confirms that we do not accidentally leak the ground truth of $h$ into the initial observation. For ``random act", a random policy is used (i.e.~uniform across actions). This fails as random actions are extremely unlikely to verify the hypothesis.

\smallsec{Discussion}
From these results, we can clearly see that naive RL and other baselines cannot efficiently solve hypothesis verification tasks. When we use our pretraining method, we use the insight that hypotheses often have a clear causal structure that can be exploited when they are formed as ``triplet hypotheses.'' Not all hypotheses fall neatly into this form, and we may not have this form for all hypotheses. But if we have some that fit this form, we can gain a foothold that lets us make progress on this problem, and then later adapt them to other hypotheses. From Figure~\ref{fig:stage280maxstd} we can see that this pretraining, triplet training and adaptation works. We also show in Section~\ref{section:intrinsic} that other plausible pretraining rewards fail to achieve the same results as our method.

Looking at the different environments, we see that in Pushblock and ColorSwitch, even with the policy learned from the triplet pre/post reward, the agent is able to generalize and perform well on hypotheses not seen in the pretraining phase as we can see in Table~\ref{table:success}. Because the finetuning step adds additional variance to training, there is a small, but non-trivial drop in average performance. This drop is less extreme when comparing max values (see Appendix~\ref{appendix:pbcs_explain}).

\begin{table}[h!]
\caption{Average Hypothesis Prediction scores, broken down by triplet (pretrained) and non-triplet (not seen in pretraining)}
\begin{center}
\begin{tabular}{ccccc}
    \multicolumn{1}{c}{ } &\multicolumn{1}{c}{\bf Method}  &\multicolumn{1}{c}{\bf Overall} &\multicolumn{1}{c}{\bf Triplet Accuracy} &\multicolumn{1}{c}{\bf Non-triplet Accuracy}
    \\ \hline 
    \multicolumn{1}{c}{\multirow{2}{*}{\bf ColorSwitch}}&Fixed Policy&86.6\%&91.1\%&82.1\%\\
    \multicolumn{1}{c}{}&Finetuned Policy&77.5\%&79.7\%&75.4\%\\
    \hline 
    \multicolumn{1}{c}{\multirow{2}{*}{\bf Pushblock}}&Fixed Policy&86.9\%&87.9\%&85.9\%\\
    \multicolumn{1}{c}{}&Finetuned Policy&85.6\%&86.3\%&84.8\%\\
    \hline 
    \multicolumn{1}{c}{\multirow{2}{*}{\bf Crafting}}&Fixed Policy&77.3\%&92.8\%&61.8\%\\
    \multicolumn{1}{c}{}&Finetuned Policy&90.7\%&98.4\%&83.0\%\\
    \hline
    \multicolumn{1}{c}{\multirow{2}{*}{\bf Cartpole}}&Fixed Policy&84.2\%&92.0\%&76.3\%\\
    \multicolumn{1}{c}{}&Finetuned Policy&92.5\%&93.4\%&91.6\%\\
    \hline
\end{tabular}
\end{center}
\label{table:success}
\end{table}

In Crafting and Cartpole on the other hand, to do well on the unseen templates, the policy also needs to be fine-tuned. This tells us that when we do have to generalize to unseen hypotheses (especially non-triplet hypotheses), adapting the policy as well as the prediction network is necessary. Recall that we test very different hypotheses such as as negations and ``independence'' hypotheses not see in triplets (see suplementary). We see from Table~\ref{table:success} that indeed, our finetuned policies greatly outperform the fixed policies on the non-triplet templates. 

We further break down these results by specific templates, entities mentioned (e.g. colors) and whether the given template was true or false. We find for instance that for most environments it is harder to determine a hypothesis is false than true. See Appendix~\ref{appendix:crosstabs} for the full analysis.

{\bf Alternate forms of pretraining}
\label{section:intrinsic}
As an ablation, we test four variants of an ``intrinsic'' reward to see if other pretraining schemes might perform equally well. We show results on the gridworld domains using 4 different intrinstic forms of pretraining: (i) change any item state in the world; receive reward at end of episode. (ii) change any item referenced in the hypothesis; receive reward at end of episode. (iii) change any item state in the world; receive reward instantaneously. (iv) change any item referenced in the hypothesis; receive reward instantaneously. Here, ``item" means any object that is not the agent (including crafting items, switches, pushblocks, etc.). (See Appendix~\ref{appendix:intrinsic}).

\begin{figure}[t]

\begin{center}
    \includegraphics[width=.8\linewidth]{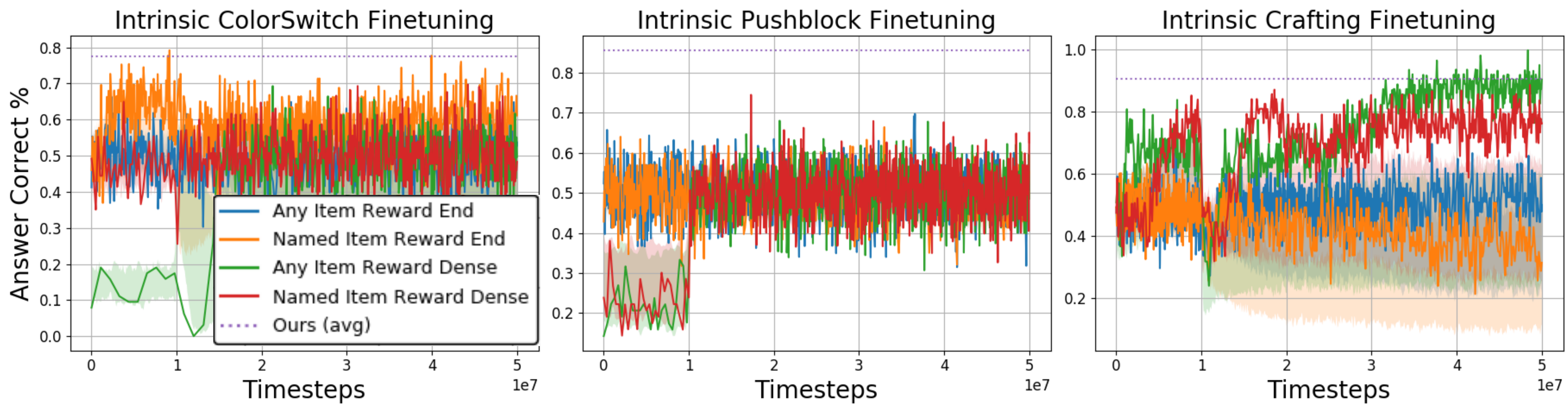}
\end{center}
\caption{Final hypothesis accuracies using alternate forms of intrinsic pretraining versus our pretraining (purple). "End" = reward only received at end of episode. "Dense" = reward received immediately after changing item state. }
\label{fig:intrinsicfinetune}
\end{figure}

In Figure~\ref{fig:intrinsicfinetune} we show the accuracies on the final hypothesis verification task for both triplet and non-triplet hypotheses, using the four intrinsic pretraining methods. We also plot the final average accuracy obtained by our adapted methods from Figure~\ref{fig:stage280maxstd}. For the intrinsic pretrained policies the best run is shown to show the best-possible case of the alternative methods. 

For Crafting the dense intrinsic pretraining works about as well as ours. This can be explained by the fact that this particular form intrinsic pretraining directly rewards the agent for doing many of the operations in the actual Crafting task, i.e.~picking up objects and crafting objects. However, averaging across the three environments, all the intrinsic pretraining methods do worse than our approach, showing the merits of our pretraining approach which exploits structure common to many hypotheses, yield an effective and general form of pretraining.

{\bf Oracle ablations}
We also show two oracle analysis in the Crafting environment. In the first, we provide an ``oracle'' hypothesis predictor which will output the ground truth of the hypothesis if it is inferable from the last $K$ frames, and test whether we can learn a policy directly using reward $R_\mathsf{Hyp}$. Similarly, we also train a hypothesis prediction network with observations from an oracle policy network (observations from which the ground truth of the hypothesis is always inferable).

\begin{figure}[h!]
\begin{center}
\begin{subfigure}{.3\textwidth}
  \includegraphics[width=1\linewidth]{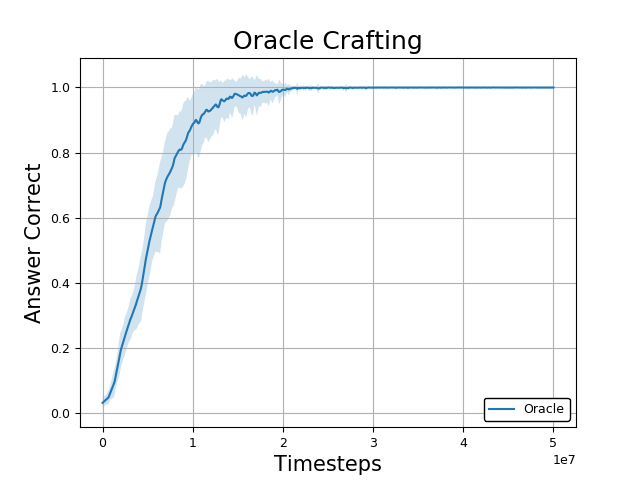}
\end{subfigure}%
\begin{subfigure}{.3\textwidth}
  \includegraphics[width=1\linewidth]{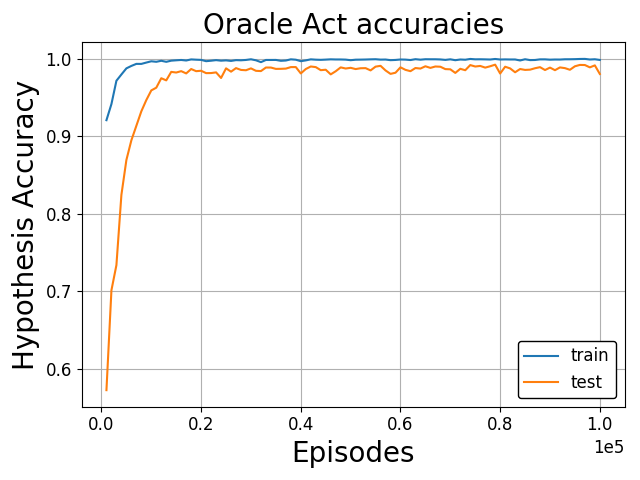}
\end{subfigure}
\end{center}
\caption{(Left) Results on training an RL using  $R_\mathsf{Hyp}$ with oracle predictor on the Crafting environment. Mean and variance on 25 random seeds are shown. (Right) Results training just hypothesis prediction on oracle policy.}
\label{fig:oracletrain}
\end{figure}
Figure~\ref{fig:oracletrain} (left) shows the prediction oracle agent quickly converging. From this we surmise that if we can predict the hypothesis already, learning an optimal policy is relatively straightforward. Similarly, (right) show that with the correct observations from an oracle policy, the hypothesis predictor is able to quickly converge as well.  This shows the joint nature of the problem (i.e. learning both policy and predictor) is what makes it challenging for the RL baseline.

\section{Discussion}
In this work, we propose a tractable formulation of the problem of training agents that can interact with a world to test hypotheses. We show that generic RL techniques struggle with the problem, but by using the common structure of some hypotheses, we are able to develop a method that works in simple environments. Specifically, we use the fact that many hypotheses can be broken into triples of the form of (pre-condition, action sequence, post-condition).We also show that once pretrained using this factorization, agents can be adapted to verify more general hypotheses.

\bibliographystyle{plain}
\bibliography{refs}

\clearpage
\appendix
\section{Templates}
~\label{appendix:templates}
\subsection{World and Hypothesis Construction}
\label{section:statements}
Returning again to our notation from the main paper, the environment at each spawn needs to construct a world $W$, and a hypothesis $h$ that is either true or false in the world. $W$ in particular describes the rules about how the environment works (i.e. which switch opens the door) which in our case can precisely be describe by a hypothesis. So given a true hypothesis, we can exactly describe the rules of the world. Therefore, in order to create an instance of a possible $W$, we can instead draw a true hypothesis about the world at random. From the hypothesis, we can then construct the rules the determine how objects in the world behave. Note that there are couple exceptions to this for our harder hypotheses, where the hypothesis can be true but only partially describes all the rules of $W$. For these cases, we draw yet another template which is consistent with the hypothesis and use that to construct the rules, such as deciding which color switch really opens the door.

Because we have to randomly give either a true or false hypothesis, we also need to be able to generate a false hypothesis for the world. So for every instance, we also draw a random false hypothesis. Now, given a true and false hypothesis, we can fully generate the world and all the items that appear in either statement. So for instance, if the true hypothesis mentions a green switch and the false one mentions a blue switch, we generate both a green and blue switch. Then, we can set the rules such that the right thing happens. So in this example, switching the green switch opens the door and the blue switch does nothing.

The final step is then to randomly choose either the true or false statement as the ``visible'' hypothesis which is passed to our agent to verify. Because we generate the world and spawn the items before we make this choice, we ensure that we do not accidentally give away the truth of the hypothesis based on what items spawned.

Our process for generating a new spawn of environment can thus be summarized as follows:

\begin{enumerate}
    \item{We randomly generate a true hypothesis}
    \item{We randomly generate a false hypothesis}
    \item{We construct a ruleset from the true hypothesis}
    \item{We spawn the agent and the items in the world described in both the true and false hypothesis}
    \item{We randomly choose either the true or false hypothesis as the ``visible'' hypothesis that the agent must verify}
\end{enumerate}

\subsection{Triplet Definitions}
\label{appendix:tripletdef}
Here we further define our terms ``pre-condition'' ``action sequence'' and ``post-condition.''

Consider a state $s$ (here we ignore the state/observation distinction for clarity of explanation, this would be an observation in our POMDP). We can consider a pre-condition or post-condition to be a boolean function of the state $s$. Given a state $s$, we can say whether a particular pre-condition or post-condition is true or false. For example, the condition ``the door is open'' is a function of the state $s$, which we can write as $f_{thedoorisopen}(s)$ which is true when the door is open, and false when it is not. Action sequences are also boolean functions, but over a sequence of states and actions. So for instance, ``the agent moves to the craftingtable'' can be written as $f_{theagentmovestothecraftingtable}(s_1, s_2, ..., a_1, a_2, ...)$ for some sequence of states and actions, which is true when the agent starts not at the craftingtable and takes actions that move it to the craftingtable.

Now we can connect these three functions or conditions to causality. In this construction, what we are doing is hypothesizing a causal connection between the pre-condition and action sequence and the post-condition: $(\text{\it pre-condition}, \text{\it action sequence})\implies \text{\it post-condition}$. This is a logical statement, which can be true or false, which asserts that when we satisfy the pre-condition and take the action sequence, the resulting state will satisfy the post-condition.

Finally, we connect this discription of these terms to the string text of the actual hypotheses. We treated these three terms like exact logical functions, but we in fact translate these functions to text strings. So a logical function of whether the door is open becomes "the door is open." In this work, we write these strings ourselves, although in future work, these could be crowd-sourced, or generated as part of an interactive game. Human language is more complicated and less un-ambigious than mathematical formulations but as we note in the main paper, we train our agents to generalize to new hypotheses and ones that do not follow this exact structure. By our choice of using templated language, we keep the kinds of hypotheses agents can learn as general as possible. Future work might explore how to solve new hypotheses online from humans, or even generate its own hypotheses to solve. 

In the next sub-section, we exhaustively list the triplet and non-triplet templates we use in this work.

\subsection{Templates}
\label{appendix:rawtemplates}

\textbf{Color Switch}: \\ \\
{\fontsize{8}{10}\selectfont
\textbf{Pre-condition}: \\
if the COLOR switch is ON\_OFF\_SWITCHSTATE \\
when the COLOR switch is in the ON\_OFF\_SWITCHSTATE position \\
the COLOR switch is ON\_OFF\_SWITCHSTATE \\ 

\textbf{Action}: \\ 
"" \\

\textbf{Post-condition}: \\ 
then the door is open \\
the door is passable \\
and we see the door is open \\
the door will open \\
        
\textbf{Finetune templates}: \\ 
the door can only be opened by switching the COLOR switch to ON\_OFF\_SWITCHSTATE \\
when we see the COLOR switch is ON\_OFF\_SWITCHSTATE the door must be open \\
if the COLOR switch turns ON\_OFF\_SWITCHSTATE the door opens \\
when we see the door open it must be that the COLOR switch is in the ON\_OFF\_SWITCHSTATE position \\
those who want to open the door must first switch the COLOR switch ON\_OFF\_SWITCHSTATE \\
no password just make the COLOR switch be ON\_OFF\_SWITCHSTATE to open the door \\
COLOR switch ON\_OFF\_SWITCHSTATE implies door is open \\
only the COLOR switch being ON\_OFF\_SWITCHSTATE opens the door \\
the door is open because COLOR switch is in the ON\_OFF\_SWITCHSTATE position \\
COLOR switch ON\_OFF\_SWITCHSTATE equals open door \\
the COLOR switch opens the door but only when it is ON\_OFF\_SWITCHSTATE \\
door is open must mean that COLOR switch is ON\_OFF\_SWITCHSTATE
an ON\_OFF\_SWITCHSTATE means the door is open but only if it is COLOR \\
COLOR controls the door and it opens when it is ON\_OFF\_SWITCHSTATE \\
ON\_OFF\_SWITCHSTATE is the correct position of the COLOR switch and it opens the door \\
the switch that causes the door to be open when it is ON\_OFF\_SWITCHSTATE is COLOR \\
if you see COLOR switch then the door is open \\
the door is independent of the COLOR switch \\
if the door is not open then the COLOR switch must be ON\_OFF\_SWITCHSTATE \\
if the COLOR switch is not ON\_OFF\_SWITCHSTATE then the door is open \\
to make the door not open the COLOR switch must be not ON\_OFF\_SWITCHSTATE \\
whether the door is open is completely independent of the COLOR switch \\
the COLOR switch is what controls the door \\
a not ON\_OFF\_SWITCHSTATE COLOR switch opens the door \\

\textbf{Template Values} \\ 
\textbf{COLOR}: \\
blue \\
red \\
green \\
black \\ \\
\textbf{ON\_OFF\_SWITCHSTATE}: \\ 
on \\ 
off \\ 
}

\textbf{Pushblock} \\ \\
{\fontsize{8}{10}\selectfont
\textbf{Pre-condition}: \\
whenever the pushblock is in the PUSHBLOCK\_POSITION \\
if the pushblock is at the PUSHBLOCK\_POSITION \\
the pushblock is at the PUSHBLOCK\_POSITION \\ 

\textbf{Action}: \\
"" \\ 
    
\textbf{Post-condition}: \\ 
then the door is open \\
the door is passable \\
and we see the door is open \\
the door will open \\ 

\textbf{SP\_FULL\_TRAIN}: \\ 
PUSHBLOCK\_POSITION is the correct position for the pushblock for the door to open \\
if the door is open it must be that the pushblock is at the PUSHBLOCK\_POSITION \\
when the door is open it is because the pushblock is in the PUSHBLOCK\_POSITION \\
when the pushblock is at the PUSHBLOCK\_POSITION the door is open \\
pushblock PUSHBLOCK\_POSITION means door open \\
the door can only be opened when the pushblock is PUSHBLOCK\_POSITION \\
if the pushblock is PUSHBLOCK\_POSITION it means the door is open \\
PUSHBLOCK\_POSITION pushblock opens the door \\
open door implies pushblock PUSHBLOCK\_POSITION \\
open door means pushblock PUSHBLOCK\_POSITION \\
door opens when PUSHBLOCK\_POSITION is where the pushblock is \\
PUSHBLOCK\_POSITION is the correct position for the pushblock to open the door \\
the door when the pushblock is PUSHBLOCK\_POSITION is open \\
PUSHBLOCK\_POSITION position of the pushblock causes the door to open \\
door only opens on PUSHBLOCK\_POSITION pushblock \\
door can only open with pushblock being PUSHBLOCK\_POSITION \\
the pushblock being at the PUSHBLOCK\_POSITION is completely independent of the door \\
the pushblock being PUSHBLOCK\_POSITION is independent of the door being open \\
the door state is independent of pushblock PUSHBLOCK\_POSITION \\
PUSHBLOCK\_POSITION pushblock and door are independent \\

\textbf{Pushblock values}: \\ 
\textbf{PUSHBLOCK\_POSITION}: \\ 
left \\ 
right \\
top \\
bottom \\ 
}

\textbf{Crafting} \\ \\
{\fontsize{8}{10}\selectfont
\textbf{Pre-condition}: \\ 
when you are at LOCATION and you have CRAFTING\_ITEM \\
you are at LOCATION and have in your inventory CRAFTING\_ITEM \\
whenever you have a CRAFTING\_ITEM and are at LOCATION \\

\textbf{Action}: \\ 
and you do CRAFTING\_ACTION \\
then you CRAFTING\_ACTION \\

\textbf{Post-condition}: \\ 
you now have CREATED\_ITEM in your inventory \\
then CREATED\_ITEM is created \\
and this creates CREATED\_ITEM \\
so CREATED\_ITEM is created and put in your inventory \\
then CREATED\_ITEM is made \\

\textbf{Finetune Templates}: \\ 
to create a CREATED\_ITEM you must have CRAFTING\_ITEM and go to LOCATION and do the action CRAFTING\_ACTION \\
CREATED\_ITEM can be created by doing CRAFTING\_ACTION at LOCATION when CRAFTING\_ITEM is in inventory \\
whenever you do CRAFTING\_ACTION and have CRAFTING\_ITEM at LOCATION a CREATED\_ITEM is made \\
you have CRAFTING\_ITEM and go to LOCATION and CRAFTING\_ACTION and CREATED\_ITEM will be created \\
whoever does CRAFTING\_ACTION at LOCATION with CRAFTING\_ITEM gets CREATED\_ITEM \\
if you have CRAFTING\_ITEM at LOCATION and you CRAFTING\_ACTION you get CREATED\_ITEM \\
if you do CRAFTING\_ACTION at LOCATION with CRAFTING\_ITEM you make CREATED\_ITEM \\
whenever you have CRAFTING\_ITEM at LOCATION and do CRAFTING\_ACTION then you make a CREATED\_ITEM \\
having CRAFTING\_ITEM in your inventory being at LOCATION and doing CRAFTING\_ACTION creates CREATED\_ITEM \\
CREATED\_ITEM can be made with CRAFTING\_ITEM when you do CRAFTING\_ACTION at LOCATION \\
CRAFTING\_ITEM plus LOCATION plus CRAFTING\_ACTION equals CREATED\_ITEM \\
create a CREATED\_ITEM by being at LOCATION with CRAFTING\_ITEM and doing CRAFTING\_ACTION \\
CRAFTING\_ACTION at LOCATION creates CREATED\_ITEM but only if you have a CRAFTING\_ITEM \\
if you want to make a CREATED\_ITEM then go to LOCATION with CRAFTING\_ITEM and do CRAFTING\_ACTION \\
CRAFTING\_ITEM in inventory at LOCATION makes CREATED\_ITEM if you do CRAFTING\_ACTION \\
CREATED\_ITEM when CRAFTING\_ITEM at LOCATION and do CRAFTING\_ACTION \\
if you are at LOCATION and do CRAFTING\_ACTION you make CREATED\_ITEM \\
if you are anywhere and do CRAFTING\_ACTION with CRAFTING\_ITEM you make a CREATED\_ITEM \\
having CRAFTING\_ITEM at LOCATION and doing CRAFTING\_ACTION does not make a CREATED\_ITEM \\
CREATED\_ITEM is created by being at LOCATION and doing CRAFTING\_ACTION \\
make a CREATED\_ITEM by having a CRAFTING\_ITEM and doing CRAFTING\_ACTION \\
you have CRAFTING\_ITEM and go to LOCATION and CRAFTING\_ACTION and CREATED\_ITEM will not be created \\
LOCATION plus CRAFTING\_ACTION creates a CREATED\_ITEM \\
with a CRAFTING\_ITEM you can make a CREATED\_ITEM by doing CRAFTING\_ACTION \\ 

\textbf{Template Values}: \\ 
\textbf{CRAFTING\_ITEM} : \\ 
iron \\
wood \\
stick \\ 
pickaxe \\ 
coal \\ \\
\textbf{CREATED\_ITEM}: \\ 
torch \\
bed \\ \\
\textbf{LOCATION}: \\ 
craftingtable \\ \\
\textbf{CRAFTING\_ACTION}: \\ 
craft \\ \\
}

\textbf{Carpole} \\ \\
{\fontsize{8}{10}\selectfont
\textbf{Pre-condition}: \\ 
when are in the COLOR zone \\
if you are in the COLOR \\
when the agent is within COLOR \\

\textbf{Action}: \\ 
"" \\

\textbf{Post-condition}: \\ 
the gravity is now MULTIPLIER \\
gravity is MULTIPLIER \\
the wind pushes DIRECTION \\
there is a DIRECTION wind \\

\textbf{Finetune Templates}: \\ 
MULTIPLIER gravity is caused by the COLOR zone \\
if you go to COLOR zone then gravity is MULTIPLIER \\
the gravity is MULTIPLIER because the agent is in the COLOR \\
COLOR zone implies MULTIPLIER gravity \\
gravity is MULTIPLIER whenever you go into the a COLOR zone \\
the COLOR causes gravity to MULTIPLIER \\
to make gravity MULTIPLIER need to be in COLOR \\
COLOR zone gravity MULTIPLIER \\
a gravity multiplier of MULTIPLIER is caused by being in COLOR zone \\
COLOR equals wind DIRECTION \\
when the wind blows DIRECTION it is because you are in COLOR \\ 
COLOR zone causes DIRECTION wind \\
only being in COLOR makes the wind blow DIRECTION \\
DIRECTION wind is in the COLOR zone \\
when you are in the COLOR zone there is a DIRECTION wind \\
DIRECTION wind is caused by being in the COLOR zone \\
wind pushing DIRECTION whenever you are in COLOR \\
gravity is totally independent of COLOR \\
COLOR zone does not effect gravity it is independent \\
the wind is completely independent of the COLOR zone \\
independent of wind DIRECTION is COLOR \\
gravity is changed by being in COLOR but not MULTIPLIER \\
the effect of being in COLOR is opposite to gravity MULTIPLIER \\
the wind blows opposite of DIRECTION when in COLOR zone \\
being in COLOR causes the wind to blow opposite to DIRECTION \\

\textbf{Template Values}: \\
\textbf{COLOR}: \\
blue \\
red \\
green \\
black \\ \\
\textbf{MULTIPLIER} : \\
decreased \\
increased \\ \\
\textbf{DIRECTION}: \\
left \\
right \\
}
\clearpage

\section{Additional Discussion / Details}
\subsection{Pretraining additional discussion}
\label{appendix:pretrainingdisc}
For the ColorSwitch environment, we found that pretraining with just the pre-condition reward leads to better results for the Color Switch environment and show those results here. We chose $C=10$ for this proxy reward, so we see from our figure that we are able to achieve the pre-condition toggling almost perfectly.

For Pushblock, we use both $R_{\mathsf{pre+post}}$ and $R_{\mathsf{post}}$, however, because of the afformentioned issue of the block getting stuck against a wall, it does not achieve perfect success each time.

For Crafting we see that pretrain results converge towards around 15. This is because the best the pretraining can do is to try to complete the recipe. If the hypothesis is false, however, we often cannot toggle the post-condition because it often requires more than $K=5$ steps to complete seperately from toggling the pre-condition. 

Similarly for Cartpole, we see the same thing, except it can get to around 17.5. This is because if the true and false zones are adjacent to each other, the agent can toggle both the pre- and post-conditions successfully (try pre-condition, when it doesn't work, move the the adjacent zone where the effect is true).

\subsection{Alternative cutoffs for adaptation training}
\label{appendix:cutoffs}
\begin{figure}[h]
\centering
\begin{subfigure}{.23\textwidth}
  \includegraphics[width=1\linewidth]{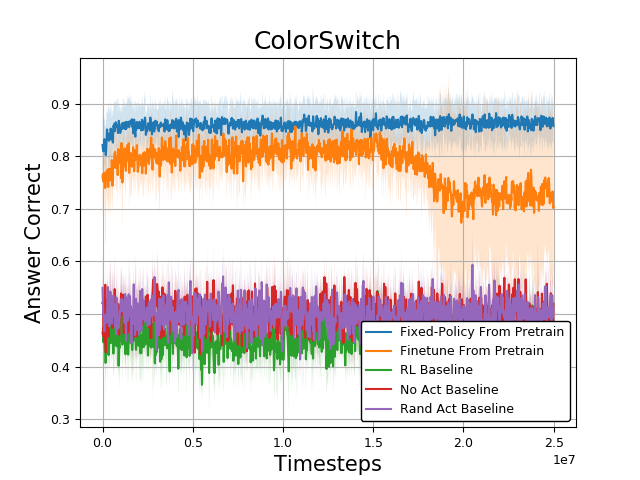}
\end{subfigure}%
\begin{subfigure}{.23\textwidth}
  \includegraphics[width=1\linewidth]{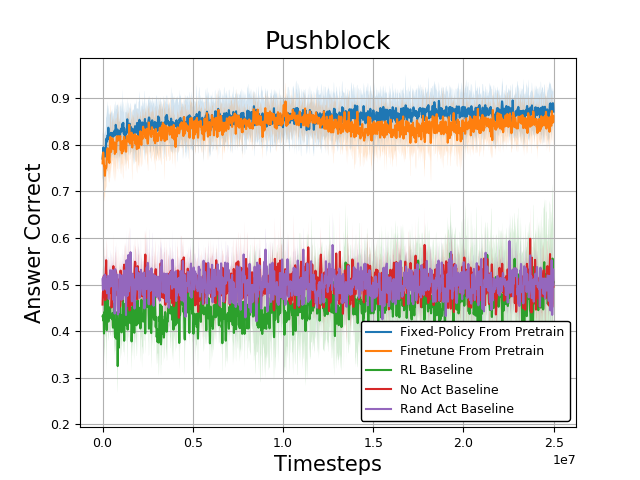}
\end{subfigure}
\begin{subfigure}{.23\textwidth}
  \includegraphics[width=1\linewidth]{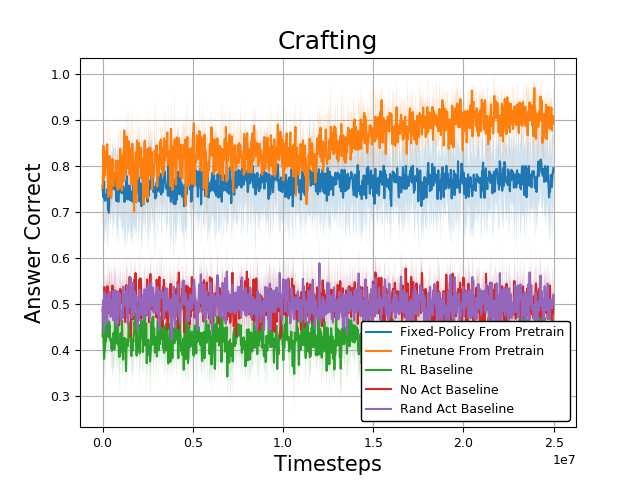}
\end{subfigure}
\begin{subfigure}{.23\textwidth}
  \includegraphics[width=1\linewidth]{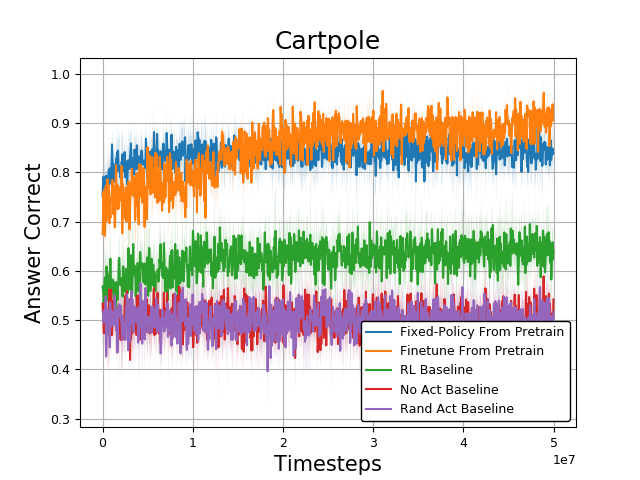}
\end{subfigure}
\caption{Hypothesis prediction accuracy on {\it both} triplet {\it and} non-triplet hypotheses for the Color Switch, Pushblock, Crafting and Cartpole environments, using $R_\mathsf{Hyp}$ reward for training.}
\label{fig:stage280maxstd_real}
\end{figure}

As promised, in Figure~\ref{fig:stage280maxstd_real} we show the adaptation results when we choose an 80\% cutoff instead of 90\%. 

\subsection{Training Variance for Pushblock and ColorSwitch}
\label{appendix:pbcs_explain}
In the main paper we discussed the problem of the finetuning training adding additional variance to training, resulting in worse average performance. First, we can actually see this from the training curves themselves in Figure 4 in the main text. The orange curves of the pretraining reward matches the blue fixed reward curve, but has more variance. In the ColorSwitch example, it has a hickup around $1.5e7$ timesteps that it does not fully recover from. This is because, unlike the fixed method, this method has to simultaneously train both the policy network and the prediction network. So when the policy network achieves a less-optimal policy, it becomes more difficult to train the prediction network which feeds back into the policy network. Whereas the fixed method only trains prediction and is much more stable during training.

We also mention that for some runs, this becomes less extreme. Some runs of the finetune policy do much better than average. In colorswitch, the best finetune run achieves $83.5\%$ accuracy versus $90.5\%$ for fixed and for pushblock achieves $87.6\%$ versus $89\%$, a smaller difference than the averages in Table 1.

\clearpage
\section{Learning Details and Hyperparameters}
\label{appendix:learningparams}
One detail of the prediction network is that we need to keep a memory of past state sequences, hypotheses and ground truths so we can actually train our prediction network. We do this by simply keeping track of the last $N$ times our agent answered a question, and keeping these in a FIFO memory. When we update our prediction network, we randomly sample from this pool. This also necessitates a 100k step break in period to collect enough examples.

In our policy finetuning experiments, we also stabilize our dual optimization problem by trading of optimization of the policy network and the prediction network. We must also start with the prediction network so that the reward for answering correctly is meaningful.

No extensive grid search of hyper-parameters was conducted. At various stages we experimented with different hyperparameter values especially PPO parameters (timesteps per batch), learning rates, PPO epochs, and optimizer. 

Experiments were performed on Tesla K40, TitanX and TitanX Pascals GPUs on 2-4 GPU workstations. Several jobs could be run on a single GPU with the greatest performance bottleneck being CPU cycles rather than GPU cycles or memory.

\begin{table}[h]
\caption{Pretraining Hyperparameters}
\label{table:pretrainhype}
\begin{center}
\begin{tabular}{ll}
\multicolumn{1}{c}{\bf Parameter}  &\multicolumn{1}{c}{\bf Value}
\\ \hline \\
Algorithm & PPO~\cite{schulman2017proximal} \\
Timesteps per batch & $2048$ \\
Clip param & $0.2$ \\
Entropy coeff & $0.1$ \\
Number of parallel processes & $8$ \\
Optimizer epochs per iteration & $4$ \\
Optimizer step size & $2.5e^{-4}$ \\
Optimizer batch size & $32$ \\
Discount $\gamma$ & $0.99$ \\ 
GAE $\lambda$ & $0.95$ \\
learning rate schedule & constant \\
Optimizer & ADAM~\cite{kingma2014adam} \\
Past Frame Window Size & $5$ \\
\end{tabular}
\end{center}
\end{table}

\begin{table}[h]
\caption{Finetuning Hyperparameters}
\label{table:finetunehype}
\begin{center}
\begin{tabular}{ll}
\multicolumn{1}{c}{\bf Parameter}  &\multicolumn{1}{c}{\bf Value}
\\ \hline \\
Algorithm & PPO~\cite{schulman2017proximal} \\
Timesteps per batch & $2048$ \\
Entropy coeff & $0.1$ \\
Number of parallel processes & $8$ \\
Optimizer epochs per iteration & $4$ \\
Optimizer step size & $1e^{-5}$ \\
Optimizer batch size & $32$ \\
Discount $\gamma$ & $0.99$ \\ 
GAE $\lambda$ & $0.95$ \\
learning rate schedule & constant \\
Optimizer & SGD \\
Past Frame Window Size & $5$ \\
\end{tabular}
\end{center}
\end{table}

\begin{table}[h]
\caption{Prediction Hyperparameters}
\label{table:predhype}
\begin{center}
\begin{tabular}{ll}
\multicolumn{1}{c}{\bf Parameter}  &\multicolumn{1}{c}{\bf Value}
\\ \hline \\
Timesteps per batch & $2048$ \\
Optimizer step size & $1e^{-3}$ \\
Optimizer batch size & $128$ \\
learning rate schedule & constant \\
Optimizer & ADAM~\cite{kingma2014adam} \\
Memory Burn-in & $100000$ \\
Memory Size & $200$ \\
Alternate Training Window & $10000000$ \\
\end{tabular}
\end{center}
\end{table}

Basis of RL implementations was from~\cite{pytorchrl}
\clearpage
\section{Network Details}
\label{appendix:networkparams}

\begin{figure}[h]
\begin{center}
   \includegraphics[width=0.75\linewidth]{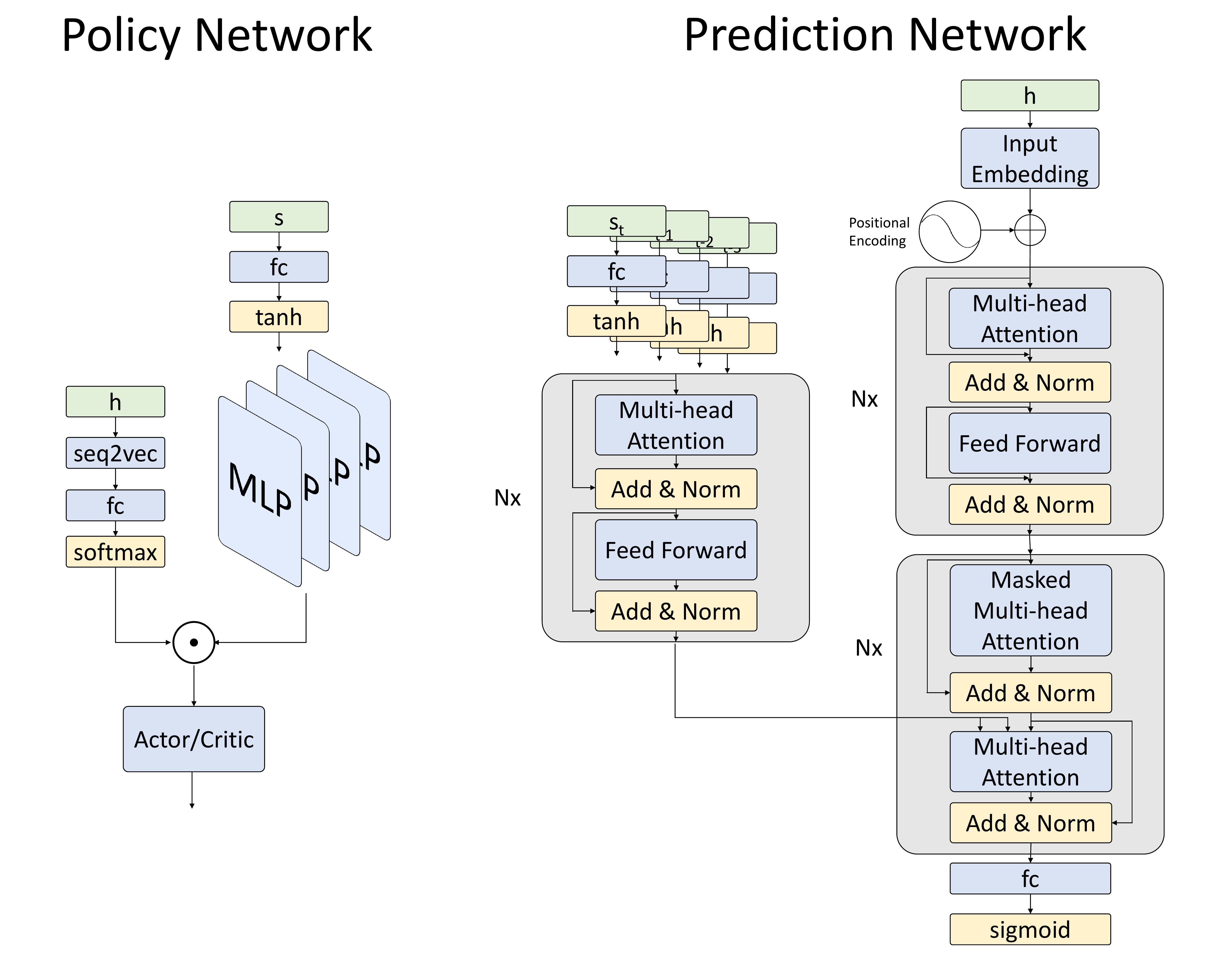}
\end{center}
\caption{Network architecture for our policy network (left) and prediction network (right)}
\label{fig:arch}
\end{figure}

Although other works such as~\cite{chaplot2018gated} have investigated language-conditioned RL (usually in the form of instruction following), our hypothesis conditioned problem proved to be challenging, and required some novelty in network architectures. Figure~\ref{fig:arch} shows our network diagrams.

Other works such as~\cite{chaplot2018gated} have incorporated gated mechanisms between language and perception. \cite{manchin2019reinforcement} employs self-attention mechanism within convolutional layers and~\cite{choi2017multi} also employs a self-attention mechanism in a DQN. Neither work incorporates language and the architectures are quite different from each other. 
Figure~\ref{fig:arch} shows the policy and transformer architectures (this is also in the main text).

For the policy network, it was important to use key-value attention.  That is: the hypothesis is fed into a seq2vec model and is used as the {\it key} of a dot-product attention mechanism. The state (the grid locations of the items in the world and the inventory of the agent if applicable) is fed as input to $N$ parallel MLPs. The output of the MLPs are then fed as the {\it values} of the attention mechanism. The output of the module is then fed into the final hidden layer of the actor-critic network. 

For the prediction network, we use the popular transformer architecture~\cite{vaswani2017attention}. Our prediction network encodes both the hypothesis and past observations (after they are passed through a one layer network) using transformer encoders. These sequences are then combined using a transformer to generate a final hidden state as output which is then fed to a final prediction layer and sigmoid function to get our binary prediction.

\subsection{Network Ablation}
\label{appendix:networkablation}
\begin{figure}[h]
\begin{center}
\begin{subfigure}{.4\textwidth}
  \includegraphics[width=1\linewidth]{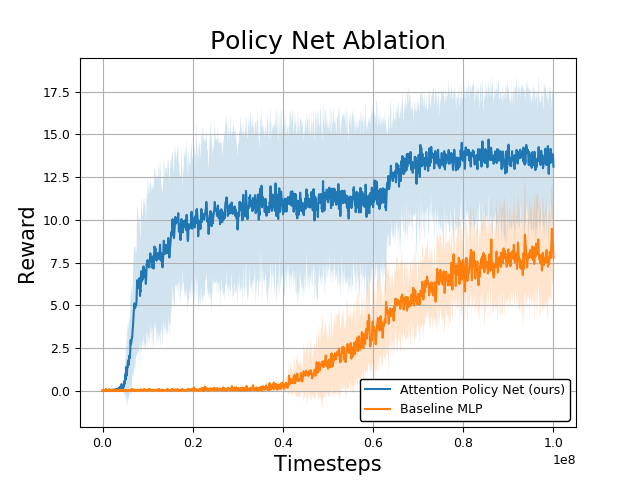}
  \label{fig:policyablation}
\end{subfigure}%
\begin{subfigure}{.4\textwidth}
  \includegraphics[width=1\linewidth]{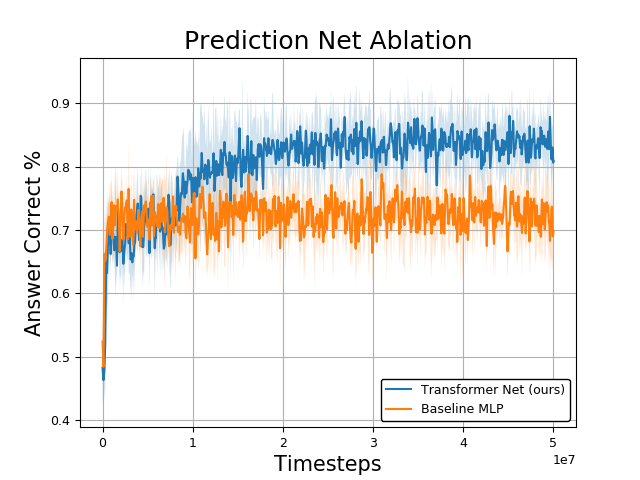}
  \label{fig:predictionablation}
\end{subfigure}
\caption{(left) policy network ablation (right) prediction network ablation.}
\label{fig:ablation}
\end{center}
\end{figure}

In Figure~\ref{fig:ablation} we see the results of our network architecture ablation. As we can see, our new policy architecture described previosuly clearly outperforms a standard MLP policy network on the language-condition pretraining task. We also see that the transformer architecture outperforms the LSTM and MLP model on the final task when we hold the policy network constant.

\subsection{Implementation and hyperparameters}
\label{appendix:implementation}
We take much of our implementation of transformers from~\cite{rush2018annotated}.

\begin{table}[h]
\caption{Policy Network Hyperparameters}
\label{table:policynethype}
\begin{center}
\begin{tabular}{ll}
\multicolumn{1}{c}{\bf Parameter}  &\multicolumn{1}{c}{\bf Value}
\\ \hline \\
Seq2Vec Model & Bag-of-Words \\
Word Embedding Size & $32$ \\
Hidden Size & $32$ \\
MLP Num Hidden Layers & $2$ \\
Number of MLP Modules & $16$ \\
Transfer Layer & $tanh$ \\
\end{tabular}
\end{center}
\end{table}

\begin{table}[h]
\caption{MLP Baseline Policy Network Hyperparameters}
\label{table:mlpbasehype}
\begin{center}
\begin{tabular}{ll}
\multicolumn{1}{c}{\bf Parameter}  &\multicolumn{1}{c}{\bf Value}
\\ \hline \\
Seq2Vec Model & Bag-of-Words \\
Word Embedding Size & $32$ \\
Hidden Size & $32$ \\
MLP Num Hidden Layers & $2$ \\
Transfer Layer & $tanh$ \\
\end{tabular}
\end{center}
\end{table}

\begin{table}[h]
\caption{Transformer Network Hyperparameters}
\label{table:txnethype}
\begin{center}
\begin{tabular}{ll}
\multicolumn{1}{c}{\bf Parameter}  &\multicolumn{1}{c}{\bf Value}
\\ \hline \\
Word Embedding Size & $32$ \\
Hidden Size & $32$ \\
Transfer Layer & $ReLU$ \\
Transformer $N$ & $3$ \\
\end{tabular}
\end{center}
\end{table}

\begin{table}[t]
\caption{Baseline Prediction Network Hyperparameters}
\label{table:blpredhype}
\begin{center}
\begin{tabular}{ll}
\multicolumn{1}{c}{\bf Parameter}  &\multicolumn{1}{c}{\bf Value}
\\ \hline \\
Seq2Vec Model & LSTM \\
LSTM Num Layers & $1$ \\
Word Embedding Size & $32$ \\
Hidden Size & $32$ \\
MLP Num Hidden Layers & $2$ \\
Transfer Layer & $tanh$ \\
\end{tabular}
\end{center}
\end{table}

\clearpage

\section{Crosstab Performance Analysis}
\label{appendix:crosstabs}
In this section, we further break down the results in the main paper by specific templates, entities mentioned (e.g. colors) and whether the given template was true or false. We do this analysis only for Fixed and Finetune as the baselines fail fairly uniformly and the analysis would be less interesting.

We get these results by running our pretrained models on at least $1000$ of each subcategory listed and averaging the results across results (and also across runs as in our main results). This gives us a good sample for analysis of which types of hypotheses our methods perform well or poorly on.

First we look at the accuracy sorting by whether the agent was given a true hypothesis to validate or a false hypothesis to invalidate in Tables~\ref{table:color_tf},\ref{table:push_tf},\ref{table:craft_tf}, and \ref{table:cart_tf}. We see that, except for finetune accuracy on crafting, the agents have a generally much easier time confirming true hypotheses.

One possible explanation for this of course is that the observations that confirm a true hypothesis and the policy needed to do this are much simpler. To confirm a true hypothesis, the agent must essentially perform the hypothesis described exactly. So if the hypothesis is that flipping the green switch on opens the door, the correct policy is exactly going to the green door and toggling. And then the state affect is immediate. If following this causes the door to open, the state changes in a very obvious way - the door opens. Whereas if the statement was false, the agent would have to look at the sequence of states and infer that indeed did exactly the right hypothesis and the effect did not happen. Noticing a change in state is often easier than noticing a lack of change.

However, there may be other subtle things going on related to the agent behavior. For instance, if an agent's policy was to guess True if it does not know the ground truth, it would by definition do better on true hypotheses. Indeed, if we look at the actions pushblock agent, we notice this. Recall that if the block is against a wall, the agent may not be able to trigger the post-condition. But actually, the agent's odds here are better than random. If the block is against a wall, it can test two possible door conditions (if it's stuck on the left, it can test up and down). By doing this, it is able to test two alternate hypotheses. If the agent wants to test a hypothesis that moving to the right opens the door, it can test moving it to up and down. If it tries both of these and the door does not open, it is more likely that it is true than not. So the agent will guess True. This leads to a high True answer rate, but this stragegy will always fail in those less likely cases where that hypothesis was actually false.

\begin{table}[h]
\caption{Colorswitch T/F Crosstabs}
\label{table:color_tf}
\begin{center}
\begin{tabular}{lll}
{\bf True/False}  & {\bf Fixed Acc} & {\bf Finetune Acc} 
\\ \hline \\
True & {\bf 0.941558} & 0.835376 \\
False & {\bf 0.788904} & 0.721134 \\
\end{tabular}
\end{center}
\end{table}

\begin{table}[h]
\caption{Pushblock T/F Crosstabs}
\label{table:push_tf}
\begin{center}
\begin{tabular}{lll}
{\bf True/False}  & {\bf Fixed Acc} & {\bf Finetune Acc}
\\ \hline \\
True & {\bf 0.969079} & 0.959164 \\
False & {\bf 0.760620} & 0.746091 \\
\end{tabular}
\end{center}
\end{table}

\begin{table}[h]
\caption{Crafting T/F Crosstabs}
\label{table:craft_tf}
\begin{center}
\begin{tabular}{lll}
{\bf True/False}  & {\bf Fixed Acc} & {\bf Finetune Acc}
\\ \hline \\
True & 0.819272 & {\bf 0.897819} \\
False & 0.730525 & {\bf 0.913886} \\
\end{tabular}
\end{center}
\end{table}

\begin{table}[h]
\caption{Cartpole T/F Crosstabs}
\label{table:cart_tf}
\begin{center}
\begin{tabular}{lll}
{\bf True/False}  & {\bf Fixed Acc} & {\bf Finetune Acc}
\\ \hline \\
True & 0.876085 & {\bf 0.894504} \\
False & 0.822460 & {\bf 0.850877} \\
\end{tabular}
\end{center}
\end{table}

Next we look at the ``elements'' i.e. the values in the templates such as the color of the switch or the item being crafted. See Tables~\ref{table:color_el},\ref{table:push_el},\ref{table:craft_el}, and \ref{table:cart_el}. With the notable exception of Cartpole, the element does not substatially change the value. The exception is Cartpole, but this is actually because unlike the other environments, there are really two distinct hypothesis effects which are not equivalent: wind versus gravity. In the hypotheses which mention left or right, the agent is actually testing hypotheses about which way the wind is blowing (left of right) whereas the half and double hypotheses are testing the gravity. We see that the agent performs better on wind hypotheses than on gravity, but the direction or gravity effect does not substantially effect the result. This is likely due to the fact that when the pole of the cartpole is level, the effect of gravity is less obvious. If you read the original cartpole code, you can see this. So these are harder because it requires the agent to have the pole in the less stable, non-centered position. Whereas for wind, the effect of a wind force is immediate and more obvious to the agent (there is now an opposing force the agent must act against to keep the pole upright).

\begin{table}[h]
\caption{Colorswitch Elements Crosstabs}
\label{table:color_el}
\begin{center}
\begin{tabular}{lll}
{\bf Element}  & {\bf Fixed Acc} & {\bf Finetune Acc}
\\ \hline \\
blue & {\bf 0.860603} & 0.817654 \\
black & {\bf 0.874475} & 0.711046 \\
red &{\bf 0.868067} & 0.699999  \\
green & 0.857772 & {\bf 0.883432} \\ 
\hline
on & {\bf 0.863618} & 0.779900 \\
off & {\bf 0.859275} & 0.772164 \\
\end{tabular}
\end{center}
\end{table}

\begin{table}[h]
\caption{Pushblock Elements Crosstabs}
\label{table:push_el}
\begin{center}
\begin{tabular}{lll}
{\bf Element}  & {\bf Fixed Acc} & {\bf Finetune Acc}
\\ \hline \\
left & {\bf 0.881174} & 0.860691 \\
bottom & {\bf 0.869612} & 0.866311 \\
right & {\bf 0.833578} & 0.830409 \\
top & {\bf 0.874413} & 0.853540 \\
\end{tabular}
\end{center}
\end{table}

\begin{table}[h]
\caption{Crafting Elements Crosstabs}
\label{table:craft_el}
\begin{center}
\begin{tabular}{lll}
{\bf Element}  & {\bf Fixed Acc} & {\bf Finetune Acc}
\\ \hline \\
pickaxe & 0.749928 & {\bf 0.885686} \\
craft & 0.774918 & {\bf 0.905880} \\
torch & 0.790433 &{\bf  0.931534} \\ 
stick & 0.790664 & {\bf 0.927625} \\
coal & 0.766912 & {\bf 0.884030} \\
bed & 0.759223 & {\bf 0.880414} \\
iron & 0.783844 & {\bf 0.890114} \\
wood & 0.808903 & {\bf 0.939032} \\
\end{tabular}
\end{center}
\end{table}

\begin{table}[h]
\caption{Cartpole Elements Crosstabs}
\label{table:cart_el}
\begin{center}
\begin{tabular}{lll}
{\bf Element} & {\bf Fixed Acc} & {\bf Finetune Acc}
\\ \hline \\
blue & 0.856136 & {\bf 0.882224} \\
black & 0.844531 & {\bf 0.868902} \\
red & 0.855209 & {\bf 0.898623} \\
green & {\bf 0.841047} & 0.840754 \\
\hline
left & 0.865710 & {\bf 0.876526} \\
right & 0.873185 & {\bf 0.901369} \\
\hline
half & 0.824625 & {\bf 0.846780} \\
double & 0.839608 & {\bf 0.846751} \\
\end{tabular}
\end{center}
\end{table}

Finally, we break out our results by each template in Tables~\ref{table:color_temp},\ref{table:push_temp},\ref{table:craft_temp}, \ref{table:craft_temp2} and \ref{table:cart_temp}. The results can be somewhat overwhelming, but a few general trends emerge. First, we can see that there is quite a bit of differing in performance for specific template types. Some specific templates can have very low performance for both fixed and finetuned. One very interesting result is for the special case templates. For Colorswitch and Pushblock, we find again that the general trend is that the fixed accuracy models perform a bit better (see Appendix~\ref{appendix:pbcs_explain}). But on Crafting and Cartpole, we see that our finetuned are able to do significantly better than fixed. In Crafting this includes negation and recipes that are missing the location or input crafting item. In Cartpole this is negation and independence statements.

\begin{table}[h]
\caption{Colorswitch Template Crosstabs}
\label{table:color_temp}
\begin{center}
\begin{tabular}{lll}
{\bf Template}  & {\bf Fixed Acc} & {\bf Finetune Acc}
\\ \hline \\
{\tiny no password just make the COLOR switch be ON\_OFF\_SWITCHSTATE to open the door} & {\bf 0.805164} & 0.715492 \\
{\tiny the COLOR switch is ON\_OFF\_SWITCHSTATE then the door is open}& {\bf 0.909462} & 0.809612 \\
{\tiny COLOR controls the door and it opens when it is ON\_OFF\_SWITCHSTATE} & {\bf 0.757894} & 0.636947 \\
{\tiny those who want to open the door must first switch the COLOR switch ON\_OFF\_SWITCHSTATE} & 0.738385 & {\bf 0.784663} \\
{\tiny COLOR switch ON\_OFF\_SWITCHSTATE equals open door} & {\bf 0.771802} & 0.654603 \\
{\tiny if the COLOR switch is ON\_OFF\_SWITCHSTATE the door is passable} & {\bf 0.909699} & 0.809354 \\
{\tiny if the COLOR switch turns ON\_OFF\_SWITCHSTATE the door opens} & {\bf 0.845888} & 0.808421 \\
{\tiny the COLOR switch is ON\_OFF\_SWITCHSTATE the door is passable} & {\bf 0.909085} & 0.806334 \\
{\tiny when the COLOR switch is in the ON\_OFF\_SWITCHSTATE position the door is passable} & {\bf 0.909311} & 0.798914 \\
{\tiny the COLOR switch is ON\_OFF\_SWITCHSTATE and we see the door is open} & {\bf 0.910084} & 0.808421 \\
{\tiny ON\_OFF\_SWITCHSTATE is the correct position of the COLOR switch and it opens the door} & {\bf 0.786958} & 0.617625 \\
{\tiny if the COLOR switch is ON\_OFF\_SWITCHSTATE the door will open} & {\bf 0.911169} & 0.813248 \\
{\tiny when we see the COLOR switch is ON\_OFF\_SWITCHSTATE the door must be open} & {\bf 0.860697} & 0.792323 \\
{\tiny if you see COLOR switch then the door is open} & {\bf 0.847181} & 0.761045 \\
{\tiny when the COLOR switch is in the ON\_OFF\_SWITCHSTATE position then the door is open} & {\bf 0.908866} & 0.805041 \\
{\tiny the COLOR switch is what controls the door} & {\bf 0.865501} & 0.790601\\
{\tiny when the COLOR switch is in the ON\_OFF\_SWITCHSTATE position the door will open} & {\bf 0.911304} & 0.801073 \\
{\tiny when the COLOR switch is in the ON\_OFF\_SWITCHSTATE position and we see the door is open} & {\bf 0.909594} & 0.800795 \\
{\tiny if the COLOR switch is ON\_OFF\_SWITCHSTATE and we see the door is open} & {\bf 0.907675} & 0.805454 \\
{\tiny the door is open because COLOR switch is in the ON\_OFF\_SWITCHSTATE position} & {\bf 0.879004} & 0.792435 \\
{\tiny the door can only be opened by switching the COLOR switch to ON\_OFF\_SWITCHSTATE} & {\bf 0.741864} & 0.727885 \\
{\tiny the switch that causes the door to be open when it is ON\_OFF\_SWITCHSTATE is COLOR} & {\bf 0.766324} & 0.701296 \\
{\tiny the COLOR switch is ON\_OFF\_SWITCHSTATE the door will open} & {\bf 0.910904} & 0.809709 \\
{\tiny if the COLOR switch is ON\_OFF\_SWITCHSTATE then the door is open} & {\bf 0.909178} & 0.803808 \\
{\tiny only the COLOR switch being ON\_OFF\_SWITCHSTATE opens the door} & {\bf 0.800769} & 0.768345 \\
{\tiny door is open must mean that COLOR switch is ON\_OFF\_SWITCHSTATE} & {\bf 0.782568} & 0.749175 \\
{\tiny an ON\_OFF\_SWITCHSTATE means the door is open but only if it is COLOR} & 0.812291 & {\bf 0.813397} \\
{\tiny when we see the door open it must be that the COLOR switch is in the ON\_OFF\_SWITCHSTATE position} & {\bf 0.827992} & 0.757434 \\
{\tiny the COLOR switch opens the door but only when it is ON\_OFF\_SWITCHSTATE} & {\bf 0.842028} & 0.685985 \\
{\tiny COLOR switch ON\_OFF\_SWITCHSTATE implies door is open} & 0.804614 & {\bf 0.806208} \\ 
\hline 
{\tiny to make the door not open the COLOR switch must be not ON\_OFF\_SWITCHSTATE} & {\bf 0.728761} & 0.698866 \\
{\tiny if the door is not open then the COLOR switch must be ON\_OFF\_SWITCHSTATE} & {\bf 0.855276} & 0.796445 \\
{\tiny a not ON\_OFF\_SWITCHSTATE COLOR switch opens the door} & {\bf 0.782377} & 0.704020 \\
{\tiny if the COLOR switch is not ON\_OFF\_SWITCHSTATE then the door is open} & {\bf 0.881558} & 0.805187 \\
{\tiny whether the door is open is completely independent of the COLOR switch} & {\bf 0.937662} & 0.850311 \\
{\tiny the door is independent of the COLOR switch} & {\bf 0.976774} & 0.803929 \\
\end{tabular}
\end{center}
\end{table}

\begin{table}[h]
\caption{Pushblock Template Crosstabs}
\label{table:push_temp}
\begin{center}
\begin{tabular}{lll}
{\bf Template}  & {\bf Fixed Acc} & {\bf Finetune Acc}
\\ \hline \\
{\tiny door can only open with pushblock being PUSHBLOCK\_POSITION} & 0.820393 & {\bf 0.856789} \\
{\tiny the pushblock is at the PUSHBLOCK\_POSITION then the door is open} & {\bf 0.869955} & 0.859727 \\
{\tiny PUSHBLOCK\_POSITION is the correct position for the pushblock to open the door} & {\bf 0.872479} & 0.843071 \\
{\tiny the pushblock is at the PUSHBLOCK\_POSITION the door is passable} & {\bf 0.872433} & 0.863410 \\
{\tiny if the pushblock is PUSHBLOCK\_POSITION it means the door is open} & {\bf 0.876048} & 0.850846 \\
{\tiny whenever the pushblock is in the PUSHBLOCK\_POSITION the door will open} & {\bf 0.874485} & 0.860658 \\
{\tiny if the pushblock is at the PUSHBLOCK\_POSITION the door is passable} & {\bf 0.874790} & 0.859502 \\
{\tiny PUSHBLOCK\_POSITION pushblock opens the door} & {\bf 0.878205} & 0.858935 \\
{\tiny the pushblock is at the PUSHBLOCK\_POSITION and we see the door is open} & {\bf 0.870191} & 0.860378 \\
{\tiny door opens when PUSHBLOCK\_POSITION is where the pushblock is} & {\bf 0.857907} & 0.853671 \\
{\tiny if the door is open it must be that the pushblock is at the PUSHBLOCK\_POSITION} & {\bf 0.868565} & 0.840094 \\
{\tiny if the pushblock is at the PUSHBLOCK\_POSITION then the door is open} & {\bf 0.877208} & 0.854677 \\
{\tiny whenever the pushblock is in the PUSHBLOCK\_POSITION the door is passable} & {\bf 0.870555} & 0.866253 \\
{\tiny open door means pushblock PUSHBLOCK\_POSITION} & 0.835129 & {\bf 0.857000} \\ 
{\tiny if the pushblock is at the PUSHBLOCK\_POSITION and we see the door is open} & {\bf 0.871365} & 0.855871 \\
{\tiny the door can only be opened when the pushblock is PUSHBLOCK\_POSITION} & {\bf 0.871308} & 0.855845 \\
{\tiny pushblock PUSHBLOCK\_POSITION means door open} & 0.839570 & {\bf 0.861030} \\ 
{\tiny whenever the pushblock is in the PUSHBLOCK\_POSITION then the door is open} & {\bf 0.878206} & 0.858112 \\
{\tiny when the door is open it is because the pushblock is in the PUSHBLOCK\_POSITION} & {\bf 0.872247} & 0.846558 \\
{\tiny whenever the pushblock is in the PUSHBLOCK\_POSITION and we see the door is open} & {\bf 0.870350} & 0.860942 \\
{\tiny open door implies pushblock PUSHBLOCK\_POSITION} & 0.832304 & {\bf 0.861506} \\ 
{\tiny if the pushblock is at the PUSHBLOCK\_POSITION the door will open} & {\bf 0.875567} & 0.856274 \\
{\tiny when the pushblock is at the PUSHBLOCK\_POSITION the door is open} & {\bf 0.878546} & 0.847757 \\
{\tiny the door when the pushblock is PUSHBLOCK\_POSITION is open} & {\bf 0.869970} & 0.849398 \\
{\tiny the pushblock is at the PUSHBLOCK\_POSITION the door will open} & {\bf 0.872346} & 0.855193 \\
{\tiny PUSHBLOCK\_POSITION position of the pushblock causes the door to open} & {\bf 0.871925} & 0.833536 \\
{\tiny door only opens on PUSHBLOCK\_POSITION pushblock} & 0.833290 & {\bf 0.843573} \\ 
{\tiny PUSHBLOCK\_POSITION is the correct position for the pushblock for the door to open} & {\bf 0.865061} & 0.846289 \\
\hline 
{\tiny the pushblock being PUSHBLOCK\_POSITION is independent of the door being open} & {\bf 0.845120} & 0.828433 \\
{\tiny the door state is independent of pushblock PUSHBLOCK\_POSITION} & {\bf 0.848979} & 0.831997 \\
{\tiny PUSHBLOCK\_POSITION pushblock and door are independent} & {\bf 0.838476} & 0.824016 \\
{\tiny the pushblock being at the PUSHBLOCK\_POSITION is completely independent of the door} & {\bf 0.851769} & 0.836789 \\
\end{tabular}
\end{center}
\end{table}

\begin{table}[h]
\caption{Crafting Template Crosstabs}
\label{table:craft_temp}
\begin{center}
\begin{tabular}{lll}
{\bf Template}  & {\bf Fixed Acc} & {\bf Finetune Acc}
\\ \hline \\
{\tiny you are at LOCATION and have in your inventory CRAFTING\_ITEM and you do CRAFTING\_ACTION then CREATED\_ITEM is made} & 0.980752 & {\bf 0.992622} \\
{\tiny if you want to make a CREATED\_ITEM then go to LOCATION with CRAFTING\_ITEM and do CRAFTING\_ACTION} & 0.618157 & {\bf 0.751656} \\
{\tiny CREATED\_ITEM when CRAFTING\_ITEM at LOCATION and do CRAFTING\_ACTION} & 0.621186 & {\bf 0.833183} \\
{\tiny you are at LOCATION and have in your inventory CRAFTING\_ITEM then you CRAFTING\_ACTION} \\ {\tiny so CREATED\_ITEM is created and put in your inventory} & 0.909046 & {\bf 0.994463} \\
{\tiny when you are at LOCATION and you have CRAFTING\_ITEM then you CRAFTING\_ACTION then CREATED\_ITEM is made} & 0.882973 & {\bf 0.990093} \\
{\tiny when you are at LOCATION and you have CRAFTING\_ITEM and you do CRAFTING\_ACTION then CREATED\_ITEM is made} & 0.988334 & {\bf 0.992137} \\
{\tiny if you do CRAFTING\_ACTION at LOCATION with CRAFTING\_ITEM you make CREATED\_ITEM} & 0.702412 & {\bf 0.981659} \\
{\tiny you are at LOCATION and have in your inventory CRAFTING\_ITEM and you do CRAFTING\_ACTION} \\ {\tiny you now have CREATED\_ITEM in your inventory} & 0.903610 & {\bf 0.993235} \\
{\tiny whenever you have a CRAFTING\_ITEM and are at LOCATION then you CRAFTING\_ACTION} \\ {\tiny you now have CREATED\_ITEM in your inventory} & 0.920456 & {\bf 0.991195} \\
{\tiny CRAFTING\_ITEM plus LOCATION plus CRAFTING\_ACTION equals CREATED\_ITEM} & 0.493720 & {\bf 0.596603} \\
{\tiny CREATED\_ITEM can be created by doing CRAFTING\_ACTION at LOCATION when  CRAFTING\_ITEM is in inventory} & 0.431615 & {\bf 0.546072} \\
{\tiny you are at LOCATION and have in your inventory CRAFTING\_ITEM then you CRAFTING\_ACTION then CREATED\_ITEM is created} & 0.980229 & {\bf 0.991235} \\
{\tiny you are at LOCATION and have in your inventory CRAFTING\_ITEM and you do CRAFTING\_ACTION then CREATED\_ITEM is created} & 0.957643 & {\bf 0.993386} \\
{\tiny CRAFTING\_ACTION at LOCATION creates CREATED\_ITEM but only if you have a CRAFTING\_ITEM} & 0.571485 & {\bf 0.913526} \\
{\tiny when you are at LOCATION and you have CRAFTING\_ITEM and you do CRAFTING\_ACTION then CREATED\_ITEM is created} & 0.988463 & {\bf 0.994632} \\
{\tiny whenever you have a CRAFTING\_ITEM and are at LOCATION then you CRAFTING\_ACTION and this creates CREATED\_ITEM} & 0.863886 & {\bf 0.990727} \\
{\tiny whenever you have CRAFTING\_ITEM at LOCATION and do CRAFTING\_ACTION then you make a CREATED\_ITEM} & 0.859223 & {\bf 0.995383} \\
{\tiny when you are at LOCATION and you have CRAFTING\_ITEM and you do CRAFTING\_ACTION and this creates CREATED\_ITEM} & 0.896466 & {\bf 0.993168} \\
{\tiny to create a CREATED\_ITEM you must have CRAFTING\_ITEM and go to LOCATION and do the action CRAFTING\_ACTION} & 0.580112 & {\bf 0.747054} \\
{\tiny whenever you have a CRAFTING\_ITEM and are at LOCATION and you do CRAFTING\_ACTION} \\ {\tiny so CREATED\_ITEM is created and put in your inventory} & {\bf 0.868949} & 0.795353 \\
{\tiny you are at LOCATION and have in your inventory CRAFTING\_ITEM then you CRAFTING\_ACTION then CREATED\_ITEM is made} & 0.935798 & {\bf 0.991869} \\
{\tiny whenever you do CRAFTING\_ACTION and have CRAFTING\_ITEM at LOCATION a CREATED\_ITEM is made} & 0.655783 & {\bf 0.918542} \\
{\tiny you are at LOCATION and have in your inventory CRAFTING\_ITEM then you CRAFTING\_ACTION and this creates CREATED\_ITEM} & 0.920180 & {\bf 0.994741} \\ 
{\tiny whenever you have a CRAFTING\_ITEM and are at LOCATION then you CRAFTING\_ACTION then CREATED\_ITEM is created} & 0.963260 & {\bf 0.993860} \\
\end{tabular}
\end{center}
\end{table}

\begin{table}[h]
\caption{Crafting Template Crosstabs (Continued)}
\label{table:craft_temp2}
\begin{center}
\begin{tabular}{lll}
{\bf Template}  & {\bf Fixed Acc} & {\bf Finetune Acc}
\\ \hline \\
{\tiny whenever you have a CRAFTING\_ITEM and are at LOCATION and you do CRAFTING\_ACTION and this creates CREATED\_ITEM} & 0.838137 & {\bf 0.994172} \\ 
{\tiny whenever you have a CRAFTING\_ITEM and are at LOCATION and you do CRAFTING\_ACTION then CREATED\_ITEM is created} & 0.891907 & {\bf 0.914471} \\ 
{\tiny you are at LOCATION and have in your inventory CRAFTING\_ITEM and you do CRAFTING\_ACTION and this creates CREATED\_ITEM} & 0.937222 & {\bf 0.990594} \\
{\tiny you are at LOCATION and have in your inventory CRAFTING\_ITEM then you CRAFTING\_ACTION} \\ {\tiny you now have CREATED\_ITEM in your inventory} & 0.885896 & {\bf 0.989596} \\
{\tiny having CRAFTING\_ITEM in your inventory being at LOCATION and doing CRAFTING\_ACTION creates CREATED\_ITEM} & 0.487408 & {\bf 0.817408} \\
{\tiny create a CREATED\_ITEM by being at LOCATION with CRAFTING\_ITEM and doing CRAFTING\_ACTION} & 0.545770 & {\bf 0.919413} \\
{\tiny when you are at LOCATION and you have CRAFTING\_ITEM and you do CRAFTING\_ACTION} \\ {\tiny so CREATED\_ITEM is created and put in your inventory} & 0.956038 & {\bf 0.991977} \\
{\tiny whoever does CRAFTING\_ACTION at LOCATION with CRAFTING\_ITEM gets CREATED\_ITEM} & 0.483942 & {\bf 0.566790} \\
{\tiny when you are at LOCATION and you have CRAFTING\_ITEM then you CRAFTING\_ACTION} \\ {\tiny so CREATED\_ITEM is created and put in your inventory} & 0.973726 & {\bf 0.996132} \\
{\tiny you have CRAFTING\_ITEM and go to LOCATION and CRAFTING\_ACTION and CREATED\_ITEM will be created} & 0.703044 & {\bf 0.914437} \\
{\tiny CRAFTING\_ITEM in inventory at LOCATION makes CREATED\_ITEM if you do CRAFTING\_ACTION} & 0.545541 & {\bf 0.897572} \\
{\tiny you are at LOCATION and have in your inventory CRAFTING\_ITEM and you do CRAFTING\_ACTION} \\ {\tiny so CREATED\_ITEM is created and put in your inventory} & 0.898226 & {\bf 0.919370} \\ 
{\tiny LOCATION plus CRAFTING\_ACTION creates a CREATED\_ITEM} & 0.541849 & {\bf 0.873599} \\
{\tiny when you are at LOCATION and you have CRAFTING\_ITEM and you do CRAFTING\_ACTION} \\ {\tiny you now have CREATED\_ITEM in your inventory} & 0.972194 & {\bf 0.994262} \\
{\tiny when you are at LOCATION and you have CRAFTING\_ITEM then you CRAFTING\_ACTION and this creates CREATED\_ITEM} & 0.941614 & {\bf 0.994693} \\
{\tiny CREATED\_ITEM can be made with CRAFTING\_ITEM when you do CRAFTING\_ACTION at LOCATION} & 0.578862 & {\bf 0.687136} \\
{\tiny whenever you have a CRAFTING\_ITEM and are at LOCATION and you do CRAFTING\_ACTION then CREATED\_ITEM is made} & 0.891969 & {\bf 0.993853} \\
{\tiny whenever you have a CRAFTING\_ITEM and are at LOCATION then you CRAFTING\_ACTION then CREATED\_ITEM is made} & 0.933055 & {\bf 0.989965} \\
{\tiny whenever you have a CRAFTING\_ITEM and are at LOCATION then you CRAFTING\_ACTION} \\ {\tiny so CREATED\_ITEM is created and put in your inventory} & 0.900469 & {\bf 0.923228} \\
{\tiny whenever you have a CRAFTING\_ITEM and are at LOCATION and you do CRAFTING\_ACTION} \\ {\tiny you now have CREATED\_ITEM in your inventory} & 0.909898 & {\bf 0.993880} \\
{\tiny when you are at LOCATION and you have CRAFTING\_ITEM then you CRAFTING\_ACTION} \\ {\tiny you now have CREATED\_ITEM in your inventory} & 0.938361 & {\bf 0.993092} \\
{\tiny if you have CRAFTING\_ITEM at LOCATION and you CRAFTING\_ACTION you get CREATED\_ITEM} & 0.832800 & {\bf 0.993352} \\
{\tiny when you are at LOCATION and you have CRAFTING\_ITEM then you CRAFTING\_ACTION then CREATED\_ITEM is created} & 0.885430 & {\bf 0.992530} \\
\hline
{\tiny you have CRAFTING\_ITEM and go to LOCATION and CRAFTING\_ACTION and CREATED\_ITEM will not be created} & 0.777171 & {\bf 0.827805} \\
{\tiny having CRAFTING\_ITEM at LOCATION and doing CRAFTING\_ACTION does not make a CREATED\_ITEM} & 0.647106 & {\bf 0.800454} \\
{\tiny make a CREATED\_ITEM by having a CRAFTING\_ITEM and doing CRAFTING\_ACTION} & 0.531967 & {\bf 0.757789} \\
{\tiny if you are anywhere and do CRAFTING\_ACTION with CRAFTING\_ITEM you make a CREATED\_ITEM} & 0.698206 & {\bf 0.875884} \\
{\tiny with a CRAFTING\_ITEM you can make a CREATED\_ITEM by doing CRAFTING\_ACTION} & 0.550868 & {\bf 0.885985} \\
{\tiny if you are at LOCATION and do CRAFTING\_ACTION you make CREATED\_ITEM} & 0.872249 & {\bf 0.999038} \\
{\tiny CREATED\_ITEM is created by being at LOCATION and doing CRAFTING\_ACTION} & 0.676661 & {\bf 0.872774} \\
\end{tabular}
\end{center}
\end{table}

\begin{table}[h]
\caption{Cartpole Template Crosstabs}
\label{table:cart_temp}
\begin{center}
\begin{tabular}{lll}
{\bf Template}  & {\bf Fixed Acc} & {\bf Finetune Acc}
\\ \hline \\
{\tiny if you are in the COLOR the wind pushes DIRECTION} & {\bf 0.93059} &
0.900575 \\
{\tiny when are in the COLOR zone gravity is MULTIPLIER} & {\bf 0.909100} & 0.859857 \\ 
{\tiny when the agent is within COLOR there is a DIRECTION wind} & {\bf 0.933811} & 0.902794 \\ 
{\tiny when the wind blows DIRECTION it is because you are in COLOR} & 0.831763 & {\bf 0.860253} \\
{\tiny a gravity multiplier of MULTIPLIER is caused by being in COLOR zone} &  0.644218 & {\bf 0.843857}  \\
{\tiny when the agent is within COLOR gravity is MULTIPLIER} & {\bf 0.912070} & 0.869424 \\
{\tiny when the agent is within COLOR the gravity is now MULTIPLIER} & {\bf 0.910071} & 0.873031 \\
{\tiny the COLOR causes gravity to MULTIPLIER} & 0.774070 & {\bf 0.835449} \\ 
{\tiny when are in the COLOR zone the gravity is now MULTIPLIER} & {\bf 0.908588} & 0.873058 \\
{\tiny COLOR zone implies MULTIPLIER gravity} & {\bf 0.802606} & 0.680320 \\ 
{\tiny if you are in the COLOR the gravity is now MULTIPLIER} & {\bf 0.909386} & 0.873600 \\ 
{\tiny gravity is MULTIPLIER whenever you go into the a COLOR zone} & 0.726433 & {\bf 0.750736}  \\
{\tiny if you are in the COLOR gravity is MULTIPLIER} & {\bf 0.910613} & 0.874824 \\
{\tiny COLOR zone causes DIRECTION wind} & 0.703039 & {\bf 0.882856} \\
{\tiny COLOR equals wind DIRECTION} & {\bf 0.848023} &  0.794995 \\ 
{\tiny when are in the COLOR zone the wind pushes DIRECTION} & {\bf 0.935098} & 0.901806 \\ 
{\tiny only being in COLOR makes the wind blow DIRECTION} & 0.739227 & {\bf 0.882116} \\ 
{\tiny if you are in the COLOR there is a DIRECTION wind} & {\bf 0.924253} & 0.895762 \\
{\tiny MULTIPLIER gravity is caused by the COLOR zone} & 0.772821 & {\bf 0.867476} \\ 
{\tiny to make gravity MULTIPLIER need to be in COLOR} & 0.735073 & {\bf 0.872627} \\
{\tiny when you are in the COLOR zone there is a DIRECTION wind} & {\bf 0.914364} & 0.892892 \\ 
{\tiny if you go to COLOR zone then gravity is MULTIPLIER} & 0.761837 & {\bf 0.826846} \\ 
{\tiny DIRECTION wind is caused by being in the COLOR zone} & 0.798327 & {\bf 0.850656} \\ 
{\tiny when are in the COLOR zone there is a DIRECTION wind} & {\bf 0.928939} & 0.902791 \\ 
{\tiny when the agent is within COLOR the wind pushes DIRECTION} & {\bf 0.930274} & 0.897901 \\ 
{\tiny the gravity is MULTIPLIER because the agent is in the COLOR} & 0.812524 & {\bf 0.852927} \\ 
{\tiny wind pushing DIRECTION whenever you are in COLOR} & 0.802038 & {\bf 0.878143} \\ 
{\tiny DIRECTION wind is in the COLOR zone} & 0.828474 & {\bf 0.899213} \\ 
{\tiny COLOR zone gravity MULTIPLIER} & {\bf 0.819835} & 0.785902
\\ \hline 
{\tiny the wind blows opposite of DIRECTION when in COLOR zone} & 0.837127 & {\bf 0.874680} \\
{\tiny the effect of being in COLOR is opposite to gravity MULTIPLIER} & 0.656581 & {\bf 0.848320} \\
{\tiny independent of wind DIRECTION is COLOR} & 0.835617 & {\bf 0.951869} \\
{\tiny gravity is totally independent of COLOR} & 0.854399 & {\bf 0.949926} \\ 
{\tiny the wind is completely independent of the COLOR zone} & 0.835538 & {\bf 0.946420} \\
{\tiny being in COLOR causes the wind to blow opposite to DIRECTION} & 0.664052 & {\bf 0.874239} \\ 
{\tiny gravity is changed by being in COLOR but not MULTIPLIER} & 0.670420 & {\bf 0.851449} \\
{\tiny COLOR zone does not effect gravity it is independent} & 0.782472 & {\bf 0.946177}\\ 
\end{tabular}
\end{center}
\end{table}

\clearpage

\section{Intrinsic Pre-training Experiments}
\label{appendix:intrinsic}
\begin{figure}[h]
\centering
\begin{subfigure}{.5\textwidth}
  \includegraphics[width=1\linewidth]{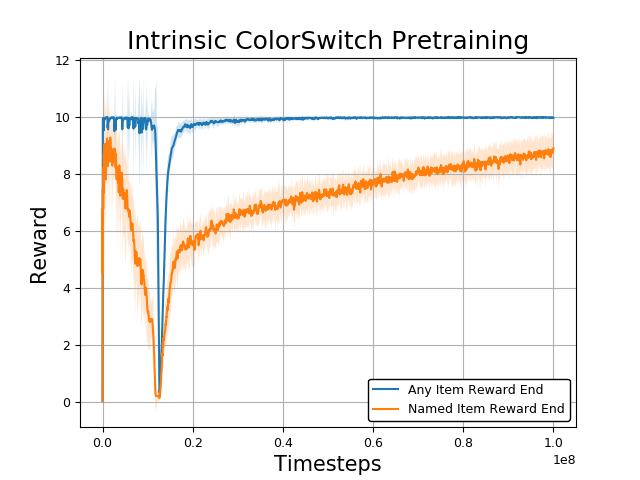}
\end{subfigure}%
\begin{subfigure}{.5\textwidth}
  \includegraphics[width=1\linewidth]{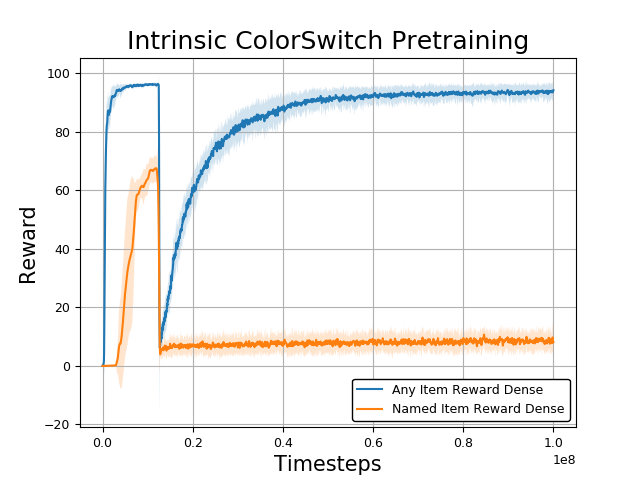}
\end{subfigure}
\caption{Pretraining Reward for ColorSwitch for intrinsic motivation. Showing mean and variance bands on 25 random seeds.}
\label{fig:intrinsiccolorswitchpretrain}
\end{figure}

\begin{figure}[h]
\centering
\begin{subfigure}{.5\textwidth}
  \includegraphics[width=1\linewidth]{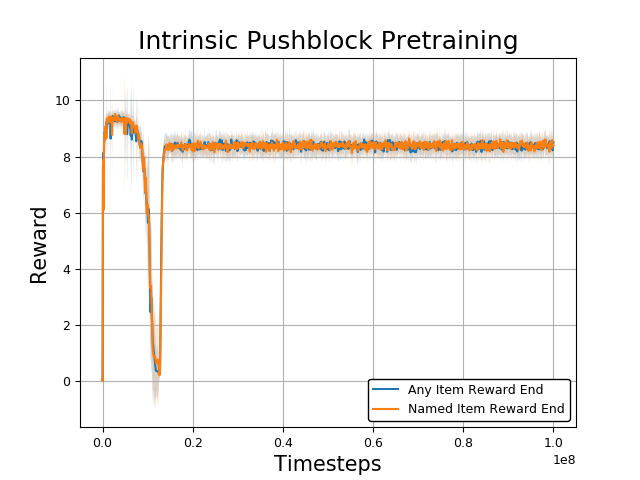}
\end{subfigure}%
\begin{subfigure}{.5\textwidth}
  \includegraphics[width=1\linewidth]{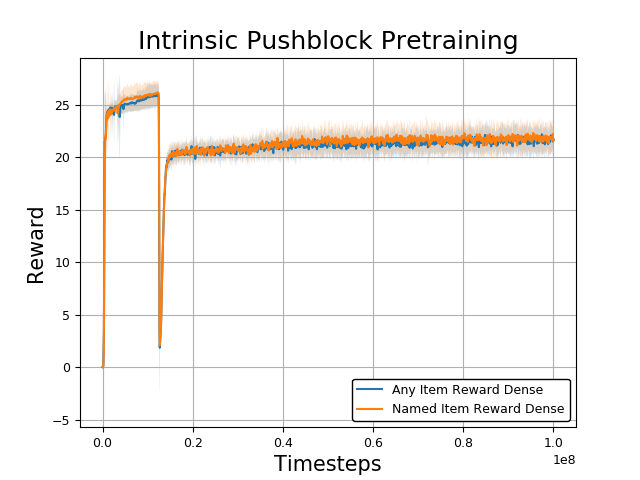}
\end{subfigure}
\caption{Pretraining Reward for Pushblock for intrinsic motivation. Showing mean and variance bands on 25 random seeds.}
\label{fig:intrinsicpushblockpretrain}
\end{figure}

\begin{figure}[h]
\centering
\begin{subfigure}{.5\textwidth}
  \includegraphics[width=1\linewidth]{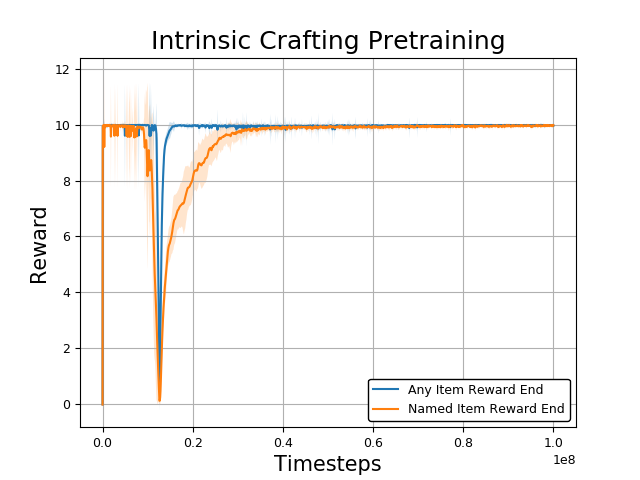}
\end{subfigure}%
\begin{subfigure}{.5\textwidth}
  \includegraphics[width=1\linewidth]{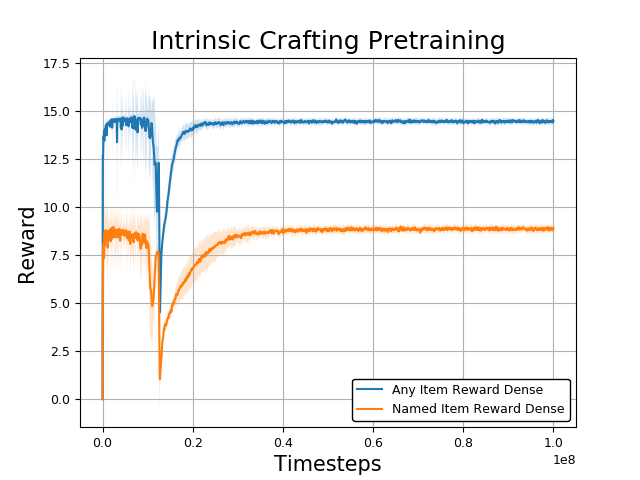}
\end{subfigure}
\caption{Pretraining Reward for Crafting for intrinsic motivation. Showing mean and variance bands on 25 random seeds.}
\label{fig:intrinsiccraftingpretrain}
\end{figure}

In this section we expand on the experiments from section ~\ref{section:intrinsic} of the main paper. 

In this experiment, we show results on our hypotheses verification problem using different forms of intrinsic motivation pre-training. We show results for 4 different pretraining schemes:
\begin{enumerate}
    \item Change any item state in the world. Receive reward at end.
    \item Change any item referenced in the hypothesis. Receive reward at end.
    \item Change any item state in the world. Receive reward instantaneously.
    \item Change any item referenced in the hypothesis. Receive reward instantaneously.
\end{enumerate}
Reward at the end means that it operates similar to our hypothesis pre-training. Specifically, the agent get reward only at the end of the episode when it has taken a stop action. At that step it gets a $+C$ reward if it changed within the last $N$ frames. For these rewards, we choose $C=10$.
Instantaneous reward is what it sounds like. When the object state is changed, the reward is instantly received by the agent. We chose $C=1$ for colorswitch and pushblock and $C=5$ for crafting.
Item means any object that is not the agent. So this includes crafting items, switches, pushblocks, etc. We show results on 25 random seeds. We preserve all training and network hyper-parameters.

We show the rest of the training curves for our intrinsic experiments in Figure~\ref{fig:intrinsiccolorswitchpretrain},~\ref{fig:intrinsicpushblockpretrain},~\ref{fig:intrinsiccraftingpretrain} we show the pretraining curves for our intrinsic rewards. 

All training and network parameters are kept the same from earlier experiments.

\end{document}